%% file: main.tex
\pdfoutput=1

\documentclass[11pt]{article}

\usepackage{acl}

\usepackage{times}
\usepackage{latexsym}

\usepackage[T1]{fontenc}

\usepackage[utf8]{inputenc}
\usepackage{comment}
\usepackage{subfigure}

\usepackage{microtype}

\input{packages}
\input{macros}
\input{todo_notes}

\newcommand{\datasetName}{K2Q\xspace}
\newcommand{\simpleDatasetName}{SD\xspace}
\newcommand{\sampledDatasetName}{mK2Q\xspace}
\title{``What is the value of \texttt{\{templates\}}?'' \\ Rethinking Document Information Extraction Datasets for LLMs}

\author{Ran Zmigrod$^*$, Pranav Shetty$^*$, Mathieu Sibue$^*$,\\ \textbf{Zhiqiang Ma, Armineh Nourbakhsh, Xiaomo Liu, Manuela Veloso} \\
        JPMorgan AI Research \\ \texttt{first.last@jpmchase.com}
}

\allowdisplaybreaks

\begin{document}
\maketitle

\def\thefootnote{*}\footnotetext{Equal Contribution.}\def\thefootnote{\arabic{footnote}}

\begin{abstract}
The rise of large language models (LLMs) for visually rich document understanding (VRDU) has kindled a need for prompt-response, document-based datasets.
As annotating new datasets from scratch is labor-intensive, the existing literature has generated prompt-response datasets from available resources using simple templates.
For the case of key information extraction (KIE), one of the most common VRDU tasks, past work has typically employed the template ``What is the value for the \texttt{\{key\}}?''.
However, given the variety of questions encountered in the wild, simple and uniform templates are insufficient for creating robust models in research and industrial contexts.
In this work, we present \datasetName,\footnote{Full dataset available upon request at \url{airdata.requests@jpmorgan.com}} a diverse collection of five datasets converted from KIE to a prompt-response format using a plethora of bespoke templates.
The questions in \datasetName can span multiple entities and be extractive or boolean.
We empirically compare the performance of seven baseline generative models on \datasetName with zero-shot prompting.
We further compare three of these models when training on \datasetName versus training on simpler templates to motivate the need of our work.
We find that creating diverse and intricate KIE questions enhances the performance and robustness of VRDU models. We hope this work encourages future studies on data quality for generative model training.


\end{abstract}

\input{sections/1intro}

\input{sections/2kie}

\input{sections/3datasets}
\input{sections/4experiments}
\input{sections/5conclusions}

\input{sections/limitations}

\input{sections/acknowledgements}

\bibliography{references}
\bibliographystyle{acl_natbib}

\clearpage
\appendix

\input{appendices/a2methodology}
\input{appendices/a1templateexamples}

\input{appendices/a3modeldetails}
\input{appendices/a4furtherexperiments}

\end{document}

%% file: packages.tex
\usepackage{enumitem}
\usepackage{relsize}
\usepackage{booktabs}
\usepackage{tabularx}
\usepackage{multirow}
\usepackage{graphicx}
\usepackage{xspace}
\usepackage{xcolor}
\usepackage{subcaption}
\usepackage{hyperref}
\usepackage[T1]{fontenc}
\usepackage{makecell}

\usepackage{amsmath}
\usepackage{amssymb}
\usepackage{amsthm}
\usepackage{amsfonts}
\usepackage{mathtools}
\usepackage{pifont}

\usepackage{tablefootnote}

\usepackage{cleveref}

\usepackage{tikz}
\usetikzlibrary{arrows.meta}
\usetikzlibrary{shapes.arrows}
\usetikzlibrary{shapes.misc}
\usetikzlibrary{fit}
\usetikzlibrary{calc,shapes}
\usetikzlibrary{tikzmark}
\usetikzlibrary{arrows}

%% file: macros.tex
\crefname{chapter}{Chapter}{}
\crefname{section}{Section}{Sections}
\crefformat{section}{Section~#2#1#3}

\crefname{table}{Table}{Tables}
\crefname{figure}{Figure}{Figures}
\crefname{algorithm}{Alg.}{Algs.}
\crefname{line}{Line}{Lines}
\crefname{appendix}{App.}{}
\crefname{chapter}{Chapter}{Chapters}

\crefname{thm}{Theorem}{Theorems}
\crefname{prop}{Proposition}{Propositions}
\crefname{definition}{Definition}{Definitions}
\crefname{lemma}{Lemma}{Lemmas}
\crefname{cor}{Corollary}{Corollaries}
\crefname{equation}{Eq.}{Eqs.}

\creflabelformat{equation}{#2\textup{#1}#3}
\crefrangelabelformat{equation}{(#3#1#4--#5#2#6)}
\creflabelformat{figure}{#2\textup{#1}#3}
\crefrangelabelformat{figure}{(#3#1#4--#5#2#6)}

\newcolumntype{L}[1]{>{\raggedright\let\newline\\\arraybackslash\hspace{0pt}}m{#1}}
\newcolumntype{C}[1]{>{\centering\let\newline\\\arraybackslash\hspace{0pt}}m{#1}}
\newcolumntype{R}[1]{>{\raggedleft\let\newline\\\arraybackslash\hspace{0pt}}m{#1}}

\renewcommand{\ldots}{\ensuremath{{\ldotp\kern-0.2em\ldotp\kern-0.2em\ldotp}}}

\renewcommand{\cdots}{\ensuremath{{\cdotp\kern-0.2em\cdotp\kern-0.2em\cdotp}}}
\renewcommand{\dots}{\ensuremath{{\ldotp\kern-0.2em\ldotp\kern-0.2em\ldotp}}}

\newcommand{\defn}[1]{\textbf{#1}}

\newcommand{\defeq}[0]{\mathrel{\stackrel{\textnormal{\tiny def}}{=}}}

\newcommand{\cmark}{\ding{51}}
\newcommand{\xmark}{\ding{55}}

%% file: todo_notes.tex
\usepackage{todonotes}
\makeatletter
\newcommand*\iftodonotes{\if@todonotes@disabled\expandafter\@secondoftwo\else\expandafter\@firstoftwo\fi}  
\makeatother
\newcommand{\noindentaftertodo}{\iftodonotes{\noindent}{}}
\newcommand{\note}[4][]{\todo[author=#2,color=#3,size=\scriptsize,fancyline,caption={},#1]{#4}} 

\newcommand{\mathieu}[2][]{\note[#1]{Mathieu}{yellow!40}{#2}}
\newcommand{\Mathieu}[2][]{\mathieu[inline,#1]{#2}\noindentaftertodo}

%% file: sections/1intro.tex
\section{Introduction}\label{sec:intro}
Visually rich document understanding (VRDU) is a core research field at the intersection of natural language processing (NLP) and computer vision.
The field aims to develop methods to process information from documents and perform natural language inference on them. VRDU is crucial for automation in industries such as finance, legal, healthcare, and government services.
Naturally, diverse and high-quality datasets are essential for training and evaluating VRDU models.
This is ever more important as generative models, such as large language models (LLMs), are becoming the mainstream state-of-the-art method for tackling VRDU problems \citep{Qwen-VL, wang2023docllm, hu-2024-docowl}.
These models require significant amounts of data for training as well as a diverse test set to evaluate whether robust document understanding is achieved.


\input{figures/flow}

VRDU covers an array of tasks that can be used to train generative models.
While visual question answering (VQA) datasets \citep{mathew2021docvqa, landeghem-23-document} can be directly used for instruction tuning, only a handful of Document VQA datasets are publicly available \citep{mathew2021docvqa, landeghem-23-document}.
Thus, existing datasets for other tasks are generally transformed into a prompt-response style to increase training volume and document diversity.

Current works that leverage existing non-VQA VRDU datasets for generative models often populate uniform templates \citep{wang2023docllm, ye-2023-ureader, tanaka-2024-instructdoc}, resulting in datasets that lack diversity and complexity.
Specifically, for key information extraction (KIE), which aims to find important entities within a document, datasets tend to populate the template ``What is the value for the \texttt{\{key\}}?'' \citep{ye-23-mplug, ye-2023-ureader, hu-2024-docowl}.\footnote{For example, finding an address on a form is converted into the question ``What is the value for the address?''}
While such questions directly translate the KIE problem, they fail to capture the complexities of VRDU and the intricacies of real-world applications of generative models for documents. Such datasets are, therefore, inadequate to train and evaluate document understanding models robustly.

In this paper, we propose a new collection of transformed KIE datasets, \defn{\datasetName}, that aims to find a balance between templated and human-annotated VQA datasets.
\datasetName is derived from five datasets: CORD \citep{park2019cord}, Docile \citep{simsa2023docile}, Kleister Charity \citep{stanislawek-2021-kleister}, and VRDU Ad-Buy (Ad-Buy) and Registration Form (Reg. Form) \citep{wang-2023-vrdu}.
We curate over $100$ different templates that lead to a diverse set of over $300,\!000$ questions across over $12,\!000$ documents.
The transformed datasets contain extractive questions as well as boolean (i.e., true or false) questions, where questions may consider multiple entities within a document.
Our key contributions 
are given below:
\begin{itemize}[noitemsep]
    \item We introduce a new publicly available collection of datasets, \datasetName, that converts five existing KIE datasets into rich and diverse prompt-response datasets.
    The creation pipeline for \datasetName is illustrated in \cref{fig:pipeline}.

    \item We show that \datasetName exhibits closer characteristics to human-made VQA datasets than simple templates through substantially lower self-BLEU and perplexity scores.\footnote{Perplexity of observing human-made questions from DocVQA and DUDE.}
    
    \item We provide zero-shot and fine-tuned benchmarks for \datasetName across seven
    models. A detailed performance breakdown is available in \cref{app:results:ablation}.


    \item We conduct an in-depth analysis of the impact of diverse, dataset-specific templates on VRDU model performance and groundedness against simple templates.
    Training on diverse templates improves relative ANLS performance by $40\%$ over simple templates.

\end{itemize}


%% file: figures/flow.tex
\definecolor{inputColor}{HTML}{648FFF}
\definecolor{questionColor}{HTML}{FE6100}
\definecolor{datasetColor}{HTML}{DC267F}
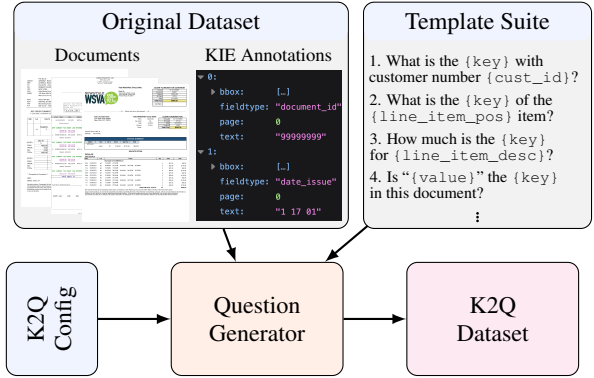
\begin{figure}[t]
    \centering
    \begin{tikzpicture}
    \node[draw, rounded corners, fill=gray!10, minimum height=3cm, minimum width=4.4cm] (og) at (1.5, -0.1) {};
    \node[inner sep=0pt] (doc1) at (0.1,-.4) {\includegraphics[width=1.4cm]{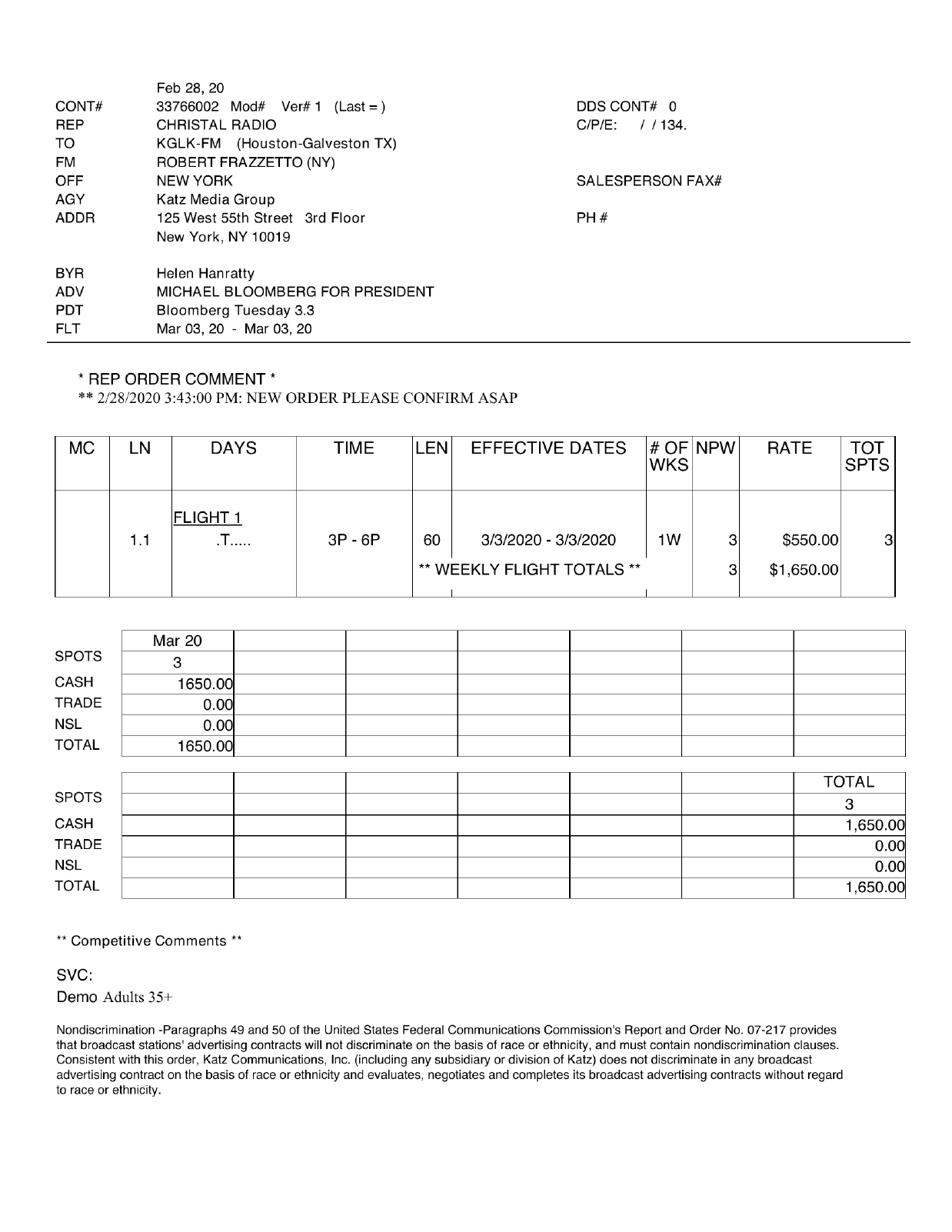}};
    \node[inner sep=0pt] (doc1) at (0.5,-0.5) {\includegraphics[width=1.4cm]{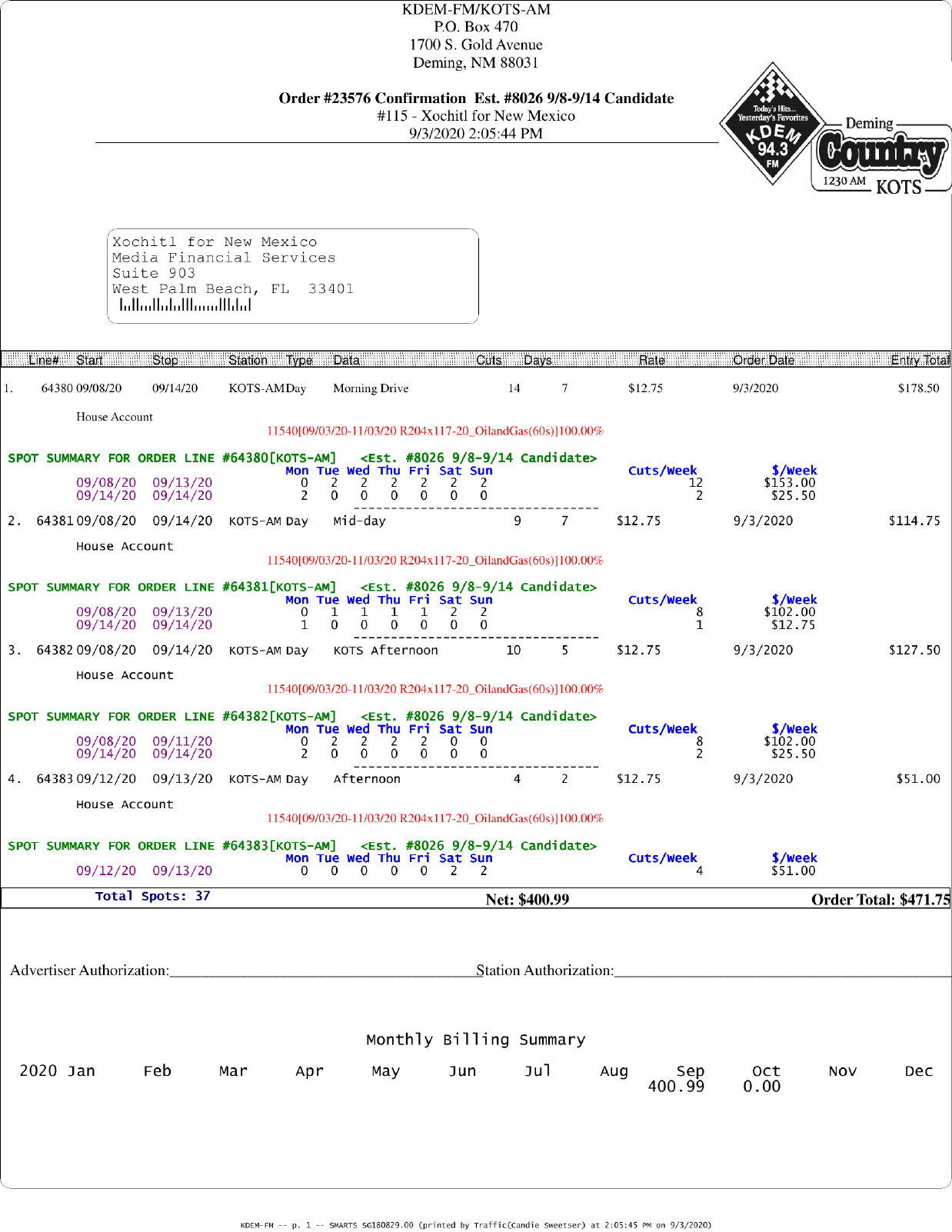}};
    \node[inner sep=0pt] (doc1) at (0.9,-0.6) {\includegraphics[width=1.4cm]{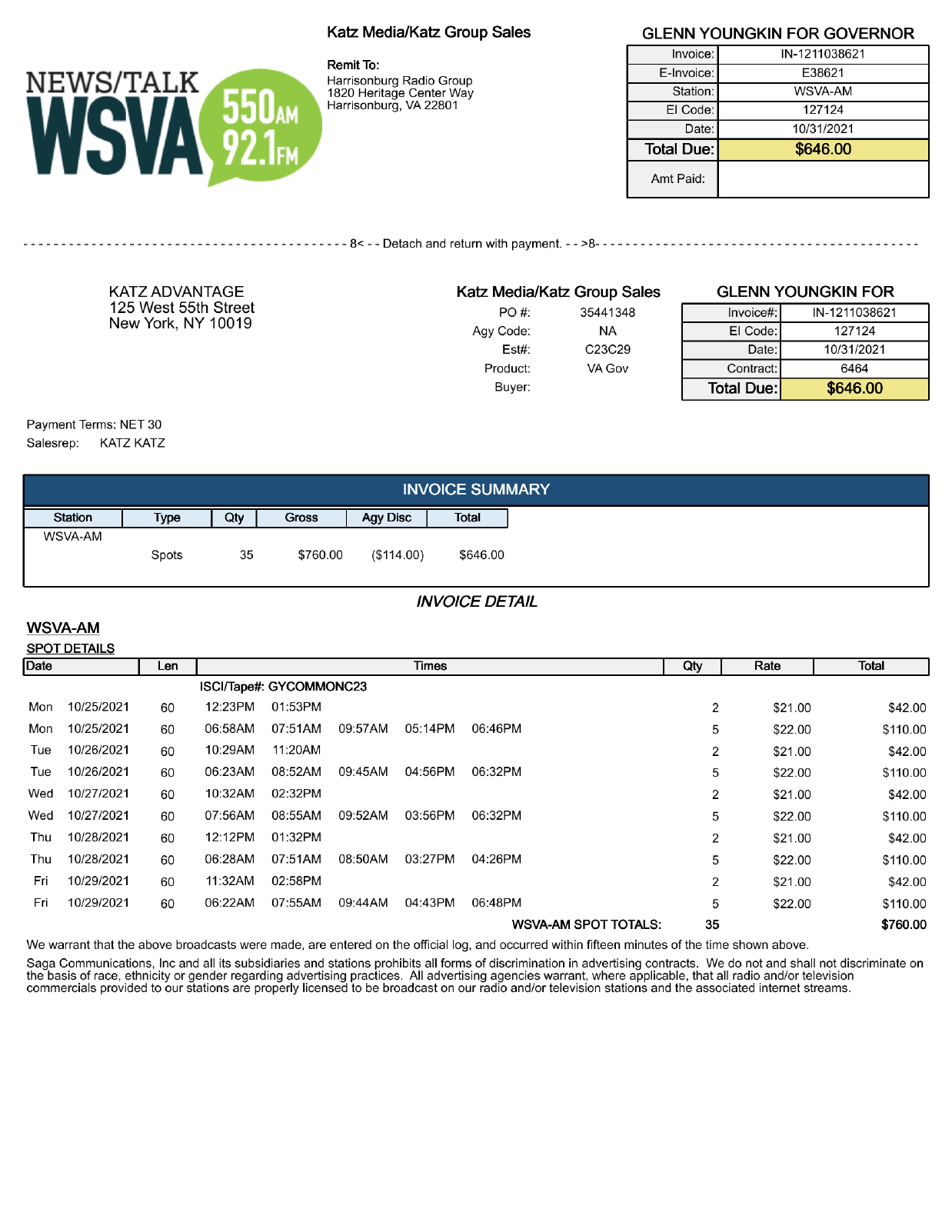}};
    \node[inner sep=0pt] (doc1) at (2.2,-0.5) {\includegraphics[height=2.cm]{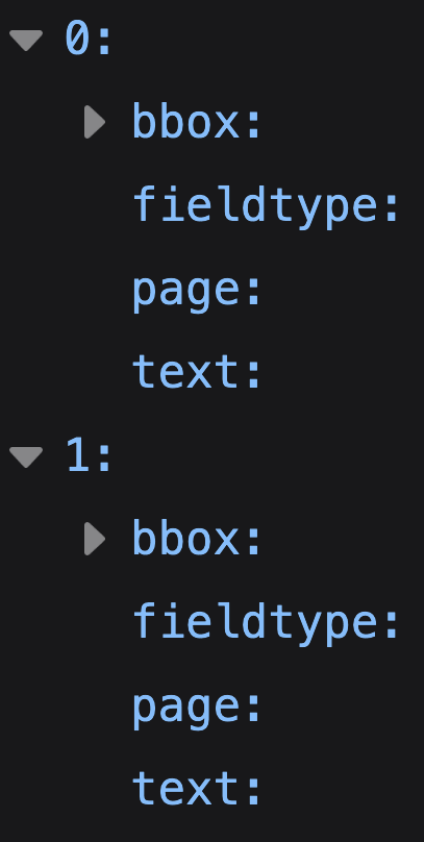}};
    \node[inner sep=0pt] (doc1) at (3.15, -0.5) {\includegraphics[height=2.cm]{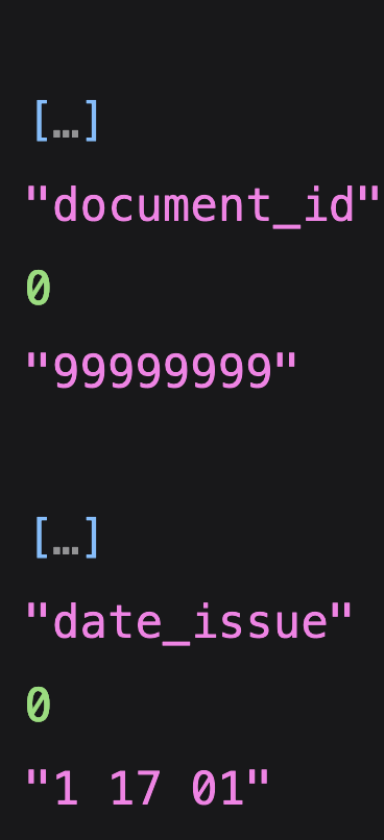}};
    \node[align=center] at (0.4, 0.7) {\scriptsize Documents};
    \node[align=center] at (2.65, 0.7) {\scriptsize KIE Annotations};
    \node[rounded corners, fill=inputColor!10, minimum height=0.5cm, minimum width=4.38cm] (background) at (1.5, 1.14) {};
    \node[align=center] at (1.5, 1.12) {\small Original  Dataset};

    \node[draw, rounded corners, fill=gray!10, minimum height=3cm, minimum width=3cm] (templates) at (5.4, -0.1) {};
    \node[rounded corners, fill=inputColor!10, minimum height=0.5cm, minimum width=2.98cm] (background) at (5.4, 1.14) {};
    \node[align=center] at (5.4, 1.12) {\small Template Suite};
    \node[align=left, font=\fontsize{6}{3.6}\selectfont] at (5.35, 0.5) {1. What is the \texttt{\{key\}} with \\ customer number \texttt{\{cust\_id\}}?};
    \node[align=left, font=\fontsize{6}{3.6}\selectfont] at (5.19, -0.01) {2. What is the \texttt{\{key\}} of the \\ \texttt{\{line\_item\_pos\}} item?};
    \node[align=left, font=\fontsize{6}{3.6}\selectfont] at (5.19, -0.53) {3. How much is the \texttt{\{key\}} \\ for \texttt{\{line\_item\_desc\}}?}; 
    \node[align=left, font=\fontsize{6}{3.6}\selectfont] at (5.23, -1.02) {4. Is ``\texttt{\{value\}}'' the \texttt{\{key\}} \\ in this document?}; 
    \node[circle,fill=black,inner sep=0pt,minimum size=1pt] at (5.4, -1.4) {};
    \node[circle,fill=black,inner sep=0pt,minimum size=1pt] at (5.4, -1.45) {};
    \node[circle,fill=black,inner sep=0pt,minimum size=1pt] at (5.4, -1.5) {};

    \node[draw, rounded corners, fill=inputColor!10, minimum height=1.2cm, minimum width=1.5cm, rotate=90, align=center] (config) at (-0.2, -2.8) {\small \datasetName \\[-.3em] \small Config};
    \node[draw, rounded corners, fill=questionColor!10, minimum height=1.5cm, minimum width=2.26cm, align=center] (questionGen) at (2.5, -2.8) {\small Question \\[-.3em] \small Generator};
    \node[draw, rounded corners, fill=datasetColor!10, minimum height=1.5cm, minimum width=2.26cm, align=center] (new) at (5.6, -2.8) {\small \datasetName \\[-.3em] \small Dataset};

    \node[] (templatePoint) at (4.09, -1.45) {};
    \draw [-latex, thick] (og) -- (questionGen);
    \draw [-latex, thick] (templatePoint) -- (questionGen);
    \draw [-latex, thick] (config) -- (questionGen);
    \draw [-latex, thick] (questionGen) -- (new);
    
    \end{tikzpicture}
    \caption{Generation pipeline of \datasetName datasets. A suite of diverse templates is designed for each specific KIE dataset. These templates are populated in accordance to a configuration file that configures the dataset size and proportion of extractive and boolean questions.}
    \label{fig:pipeline}

    \vspace{-1em}
\end{figure}

%% file: sections/2kie.tex
\section{Key Information Extraction in VRDU}
We focus on KIE in this work due to the abundance of existing datasets in the VRDU literature and the importance of information extraction in industry. We hope future work will examine other popular structured VRDU tasks such as document classification \citep{harley2015icdar} or document structure prediction \citep{li2020docbank, xu-etal-2022-xfund, zmigrod-2024-treeform}.
KIE is the task of identifying important entities within a document.
As such, many datasets have naturally examined invoices and receipts \citep{huang-2019-icdar, park2019cord, sun-2021-spatial, simsa2023docile}, where entities are clearly defined and often connect through \defn{line items} (e.g., the name and price of an item appear on the same ``line'').
Similarly, forms and legal documents have also been the focus of a plethora of datasets \citep{jaume2019funsd, Borchmann2021DUEED, stanislawek-2021-kleister, wang-2023-vrdu, zmigrod-2024-buddie}.
We provide details regarding available KIE datasets in \cref{tab:kie-datasets}.\footnote{We only focus on English datasets in this work. Nevertheless, other non-English VRDU datasets exist \citep{wang-2021-towards, qi-2022-dureadervis, ding-2023-form}.}

\input{tables/kie_datasets}

\subsection{KIE for Generative Models}\label{sec:kie:related}
\input{figures/example_questions}

To train generative models for KIE, datasets must be formulated into a prompt-response type structure.
So far, the VRDU and NLP literature has proposed three main strategies to obtain such formulations: populating simple templates based on existing annotations, manually curating questions, or generating questions using LLMs.
We discuss the advantages and disadvantages of each method below to motivate our approach in \cref{sec:dataset}.

\paragraph{Populating Simple Templates.}
The most common method for designing large datasets of instructions for LLMs is templating.
This is the case in both general NLP \cite{chung-22-scaling, wei-22-finetuned} as well as in VRDU \cite{tang-23-unifying, ye-2023-ureader, zhang-2023-llavar, wang-2023-vrdu, tanaka-2024-instructdoc, zmigrod-2024-buddie}.
Unfortunately, most templating approaches in VRDU rely on simplistic and dataset-agnostic templates that do not reflect a model's true understanding of the task nor the real-life queries that would be submitted to a generative model.
For instance, \citet{ye-23-mplug, ye-2023-ureader, wang2023docllm, hu-2024-docowl} invoke a single simple template: ``What is the value for the \texttt{\{key}\}?''.
This can lead to under-specified questions, which our work aims to avoid. 
BuDDIE \citep{zmigrod-2024-buddie} is similar to these works but also includes boolean questions.
In comparison, our work populates much richer templates thanks to the inclusion of multiple rephrasings per question, multiple entities per question, and multiple question types.
Another approach to templating is that of the InstructDoc (ID) collection \citep{tanaka-2024-instructdoc}, which features five rephrased templates per dataset that ask the model to classify text snippets into one of the key entity types.
Unfortunately, we believe this to be unrepresentative of the KIE task in practice, as the underlying goal of KIE is to \textit{directly extract} entities from a large amount of text instead of simply classifying candidate extractions.
Therefore, we do not include such templates here.

\paragraph{Manually Constructing Questions.}
Rather than relying on existing datasets from other tasks, researchers have also
constructed prompt-response datasets from the ground up, such as DocVQA \citep{mathew2021docvqa, tito-2022-hierarchical} and DUDE \citep{landeghem-23-document}. Due to the lack of other VQA datasets truly focused on visually rich documents, the literature has also considered datasets from adjacent areas such as InfographicsVQA \citep{mathew-2022-info} or ChartQA \citep{masry-2022-chart}.
The benefits of human-annotated datasets come from the quality, diversity, and depth of the questions that can be achieved.
Unfortunately, curating VQA datasets is expensive, and thus, creating datasets of the scale needed to instruction-tune LLMs can be inaccessible to many practitioners.
Our work addresses this by applying human intervention at the dataset level rather than at the document level.
We use annotators to manually design templates, resulting in a collection of datasets over 
$6$ and $7$ times larger than DocVQA and DUDE respectively.\footnote{This size factor is after we sample our selection of templates. One can increase the sampling ratio to make a significantly larger dataset.}

\input{tables/dataset_stats}

\paragraph{Generating Questions using LLMs.}
Past work has demonstrated that utilizing LLMs to create fine-tuning datasets may be a promising research avenue
\citep{honovich-2023-unnatural, peng-2023-instruction, cheung-etal-2024, alpaca, wang-23-self}.
This also includes work for general multi-modal LLMs \citep{haotian-2023-visual, li-2023-blip}, suggesting this methodology can extend to VRDU.
However, to the best of our knowledge, the VRDU literature has not investigated the use of LLMs to generate prompt-response pairs yet.
While we do not leverage LLMs to generate data in this work, we believe a future iteration of \datasetName could use an LLM to 
augment our set of
templates.



\subsection{Modeling Spectrum for KIE}
Models for VRDU (and therefore KIE) lie on a spectrum of those that only utilize text extracted from documents through optical character recognition (OCR) \citep{devlin-2019-bert, liu-2019-roberta} to those that ingest documents directly as image inputs \citep{lee-2023-pix2struct}.
On this spectrum, some models are more focused on OCR and spatial features \citep{xu2020layoutlm, peng-2022-ernie}, while others are more focused on complementing a main visual channel with text \citep{kim-2022-ocr, davis-2022-end, tang-23-unifying}.
Multi-modal LLMs have also been developed along this spectrum \citep{wang2023docllm, ye-2023-ureader, ye-23-mplug, hu-2024-docowl, tanaka-2024-instructdoc}.

%% file: tables/kie_datasets.tex
\begin{table}[t]
    \centering
    \small 
    \begin{tabular}{l rrrc}
        \bf Dataset & \bf Docs & \bf Pages & \bf Labels & \bf Lines \\
        \midrule
        Ad-Buy & $641$ & $1,\!598$ & $14$ & \cmark \\
        CORD & $1,\!000$ & $1,\!000$ & $33$ & \cmark \\
        Docile & $5,\!680$ & $7,\!372$ & $55$ & \cmark  \\
        KLC  & $2,\!778$ & $62,\!010$ & $8$ & \xmark \\
        Reg. Form & $1,\!915$ & $3,\!890$ & $6$ & \xmark \\ \midrule
        BuDDIE & $1,\!665$ & $1,\!665$ & $69$ & \xmark \\
        DeepForm & $1,\!100$ & $4,\!720$ & $5$ & \xmark \\
        FUNSD & $199$ & $199$ & $4$ & \xmark \\
        KNDA & $540$ & $3,\!229$ & $4$ & \xmark \\
        SROIE & $600$ & $1,\!000$ & $4$ & \xmark \\
        WildReceipt & $1,\!786$ & $1,\!786$ & $25$ & \xmark
        
    \end{tabular}
    \caption{KIE Datasets. The first section contains the datasets we transform in \datasetName. The second section contains other relevent KIE datasets. The \emph{Lines} column indicates whether the dataset contains line items.}
    \label{tab:kie-datasets}

    \vspace{-1em}
\end{table}

%% file: figures/example_questions.tex
\definecolor{adbuy}{HTML}{648FFF}
\definecolor{cord}{HTML}{785EF0}
\definecolor{docile}{HTML}{DC267F}
\definecolor{klc}{HTML}{FE6100}
\definecolor{regform}{HTML}{FFB000}

\begin{figure*}[t]
    \centering
    \begin{tikzpicture}
    \node[rounded corners, fill=gray!10, minimum height=6cm, minimum width=16cm] (background) at (1.5, -0.1) {};
    \node[draw=adbuy, rounded corners, fill=adbuy!10, minimum height=1cm, minimum width=14cm] (background) at (2.35, 2.2) {};
    \node[draw=cord, rounded corners, fill=cord!10, minimum height=1cm, minimum width=14cm] (background) at (2.35, 1.05) {};
    \node[draw=docile, rounded corners, fill=docile!10, minimum height=1cm, minimum width=14cm] (background) at (2.35, -0.1) {}; 
    \node[draw=klc, rounded corners, fill=klc!10, minimum height=1cm, minimum width=14cm] (background) at (2.35, -1.25) {};
    \node[draw=regform, rounded corners, fill=regform!10, minimum height=1cm, minimum width=14cm] (background) at (2.35, -2.4) {};

    \node[] at (-5.4, 2.2) {Ad-Buy};
    \node[] at (-5.3, 1.05) {CORD};
    \node[] at (-5.3, -0.1) {Docile}; 
    \node[] at (-5.2, -1.25) {KLC}; 
    \node[] at (-5.6, -2.4) {Reg. Form}; 
    
    \node[] at (2.3, 2.4) {\small \textsc{Q}: When does the advertisement start running on channel KPTH on program ``People's Court''? \textsc{A}: 01/30/20};
    \node[] at (0.4, 2.) {\small \textsc{Q}: Did the advertisement cost Teresa Tomlinson for Senate \$1,855.00? \textsc{A}: Yes};

    \node[] at (-0.34, 1.25) {\small \textsc{Q}: How much did the order(s) of Combo 1 cost in total? \textsc{A}: 30.000};
    \node[] at (0.8, 0.85) {\small \textsc{Q}: Is "20.000" the number of servings/quantity of a menu item in this receipt? \textsc{A}: No}; 

    \node[] at (-0.1, 0.1) {\small \textsc{Q}: How much is the total amount with tax of the 19th item? \textsc{A}: 130.00};
    \node[] at (1.67, -0.3) {\small \textsc{Q}: Is ``Alyse For Alaska'' the name of the customer that is being invoiced in this document? \textsc{A}: No};

    \node[] at (1.9, -1.05) {\small \textsc{Q}: How much did the charity with number 1154288 earn annually in British pounds? \textsc{A}: 36493079.00};
    \node[] at (.72, -1.45) {\small \textsc{Q}: Is 287408.00 pounds the annual expenditure of Campaign For Learning? \textsc{A}: Yes}; 

    \node[] at (-0.77, -2.2) {\small \textsc{Q}: Who is the agent of the form? \textsc{A}: Hogan \& Hartson LLP};
    \node[] at (-1.25, -2.6) {\small \textsc{Q}: Is manager the title of James A. Coppola? \textsc{A}: No};

    \end{tikzpicture}
    \caption{Examples of populated questions and answers from \datasetName.}
    \label{fig:example-questions}

    \vspace{-1em}
\end{figure*}

%% file: tables/dataset_stats.tex
\begin{table*}[t]
    \centering
    \small
    \begin{tabular}{
    L{1.4cm}
    C{.8cm}
    C{1.1cm}
    C{1.cm}
    C{1.3cm}
    C{1.0cm}
    C{1.1cm}
    C{1.2cm}
    C{1.2cm}
    }
        \multicolumn{1}{C{1.4cm}}{\bf Dataset Name} & 
        \multicolumn{1}{C{.8cm}}{\bf Num. Temp.} & 
        \multicolumn{1}{C{1.1cm}}{\bf Num. Ques.} & 
        \multicolumn{1}{C{1.cm}}{\bf \% Extr. Ques.} & 
        \multicolumn{1}{C{1.3cm}}{\bf \% 1-Page Ques.} &
        \multicolumn{1}{C{1.0cm}}{\bf Ques. Length} & 
        \multicolumn{1}{C{1.1cm}}{\bf Answer Length} & 
        \multicolumn{1}{C{1.2cm}}{\bf Ent. per Ques.}
        & 
        \multicolumn{1}{C{1.2cm}}{\bf Ques. per Doc.}
        \\ \midrule
        Ad-Buy & $50$ & $15,\!119$ & $57.8$ & $96.8$ & $9.5$ & $2.3$ & $1.5$ & $23.6 $\\
        CORD & $22$ & $39,\!575$ & $49.6$ & $100.0$ & $12.6$ & $1.4$ & $1.4$ & $39.6$ \\
        Docile & $17$ & $185,\!557$ & $68.1$ & $99.9$ & $11.6$ & $3.6$ & $1.1$ & $32.7$ \\
        KLC & $31$ & $44,\!813$ & $43.2$ & $88.3$ & $9.8$ & $1.6$ & $1.8$ & $16.1$ \\
        Reg. Form & $18$ & $23,\!427$ & $37.7$ & $99.9$ & $8.8$ & $3.1$ & $1.1$ & $12.2$
    \end{tabular}
    \caption{\datasetName statistics per dataset. Question length and answer length show the average number of tokens of questions and answers respectively. ``Extr.'' stands for extractive and ``Ent.'' for entities.}
    \label{tab:dataset-stats}

    \vspace{-1.3em}
\end{table*}

%% file: sections/3datasets.tex
\section{The \datasetName Dataset}\label{sec:dataset}
We present \defn{K}ey Information Extraction Transformed to Visual \defn{Q}uestion Answering Datasets, or \defn{K2Q} for short.
This is a collection of five KIE datasets transformed into prompt-response type datasets (similar to VQA): Ad-Buy, CORD, Docile, Kleister Charity, and Reg. Form.
These datasets were chosen 
as they cover
a wide range of domains within the field of VRDU.
\datasetName
 contains more than $300,\!000$ questions over $12,\!000$ documents.
\cref{fig:example-questions} provides
several examples from \datasetName along with their respective source dataset.
The questions are designed to reflect the intricacies of document understanding, such as co-references 
(e.g., ``How much did \textit{the charity with number 1154288} earn annually in British pounds?''), disambiguation of similar entities 
(e.g., ``How much is the total amount with tax of \textit{the 19th item}?''), and questions spanning multiple pages.
\cref{tab:dataset-stats} shows a statistical breakdown of the questions generated for \datasetName. Data split sizes are also provided in \cref{app:dataset:split}.

\subsection{Dataset Construction}
Unlike other template-based approaches, we manually construct several templates \textit{at the dataset level} for each entity in the KIE topology. Templates were created by three VRDU researchers who were already familiar with the datasets. The datasets were split evenly among them for template creation and validation.
When possible, questions are created in relation to other entities as shown in \cref{fig:example-questions} (e.g., ``Is manager the title of James A. Coppola?'' involves two distinct key entities from Reg. Form). 
In particular, CORD, Docile, and Ad-Buy contain line items that connect several different entities in an entry (similar to the rows of a table).
Examples of such items are lines in a receipt and advertisement slots. For single-page questions in \datasetName, we ensure that all entities in the question and answer are present in the input page fed to models.
Furthermore, \datasetName contains \defn{extractive} and \defn{boolean} questions; extractive questions have answers that are key entity values, while boolean questions have answers that are ``Yes'' or ``No''.
While \datasetName is intended for training generative models, we ensure it can be used as a VQA training dataset for traditional VRDU models such as \citet{xu2020layoutlm} by providing OCR token span annotations. This required heuristics for Docile and KLC, which do not relate KIE entities to the OCR.
To ensure the quality of the questions generated, we address several complexities.

\paragraph{Cleaning OCR Output.}
In the traditional KIE task, predictions are typically given as OCR token spans, which can be noisy.
Thus, cleaning the OCR entity values is necessary to enable generative models trained on \datasetName to produce more natural responses. 
We provide more details regarding our data cleaning in \cref{app:dataset:clean}.
Note that the OCR output provided in the original datasets is used to train and test the OCR-based models in \cref{sec:experiments}.

\paragraph{Under-specified Questions.}
It is sometimes possible for a question to have several correct answers within a document.
\datasetName handles such cases in two different ways.
Firstly, if a document contains multiple entities of the same type that are not line items (e.g., several vendor names mentioned), we consider all entities to be allowed as an answer. This under-specificity only occurs for question-answer pairs from Docile.
Secondly, if multiple line item rows contain repeated information, we avoid asking questions that may cause ambiguity when determining the row to consider.
For example, if a receipt contains two items that both cost $\$10$, we do not include the question ``Which item in the receipt cost $\$10$?'' as it is unclear which line item is being referred to. While it may be desirable for models to learn to return all relevant line items in this case (multi-span answer), it complicates training and evaluation, so we omit these questions. 
Future work will look into incorporating such questions.

\paragraph{Negative Boolean Question Generation.}
Half of all boolean questions in \datasetName are designed to be false.
To achieve that, we devise a candidate set of false answers 
by drawing inspiration from
the approach introduced in \citet{zmigrod-2024-buddie}.
Firstly, we consider all unique values of a given entity type in the dataset.
Secondly, for datasets with a hierarchical entity ontology, we consider unique values in the same document that share a parent entity.
Lastly, we also consider other document values with the same format (e.g., string, integer, date, etc.).
The false candidate is then sampled from one of these three sets.

\input{tables/dataset_properties}

\paragraph{Question Sampling.} 
Due to the nature of templates, we do not generate all possible questions for each document to avoid introducing too much redundancy. 
For Ad-Buy, Kleister Charity, and Reg. Form, this was achieved by randomly sampling one extractive question per entity and then a fixed number of boolean questions.
Due to the high number of entities for CORD and Docile, we generate all questions and sample after generation to select a fixed number of questions in both the extractive and boolean settings.


\input{tables/validation}

\paragraph{Dataset Validation.}
\datasetName is generated systematically after manually curating templates; nevertheless, we applied various forms of human validation to ensure a \textit{high-quality} dataset.
Prior to question generation, at least one other template writer reviewed every template to verify grammatical correctness and variety.
Post question generation, five documents from each dataset were randomly sampled.
For each document, up to ten extractive and ten boolean questions are sampled, and three validators check questions for grammatical correctness and other data issues.
An issue was considered if two out of three validators noted it.
The validators were provided with guidelines and performed the validation exercises independently. The error types and the guidelines provided to validators are discussed in more detail in \cref{app:dataset:errors}.
The overall Fleiss-kappa scores indicate moderate agreement between raters (\cref{tab:validation}). The disagreements demonstrate the complexity of assessing erroneous questions. For instance, cleaning errors may be subjective, as validators may disagree on whether or not specific words should be capitalized. Nevertheless, the low percentage of cleaning errors observed indicates this is not an issue.
We also note that half of the errors are due to annotation issues with the original datasets, and so inevitably occur in \datasetName.

\input{tables/dataset_comparison}

\subsection{Comparison of the Core Characteristics of \datasetName and Related Datasets}\label{sec:dataset:simple}
Admittedly, \datasetName requires more care and effort to collate than past template-based datasets.
We empirically motivate why this additional work is worthwhile by comparing intrinsic and extrinsic characteristics of \datasetName against those of simpler template approaches. 
This section analyzes the advantages of \datasetName in terms of data volume, diversity, and resemblance to human data.
\cref{sec:experiments} delves into the benefits of \datasetName for model training and evaluation. 

To facilitate the comparison between \datasetName and previous work \citep{hu-2024-docowl, ye-2023-ureader, wang2023docllm}, we construct baseline datasets where we use the template ``What is the value for the \texttt{\{key\}}?'' for all entities that are not line items.\footnote{We do not create line item questions with this template as they would be ambiguous for models. For example, if a receipt has multiple items, the question ``What is the value for the receipt item?'' is too ambiguous to answer.
This can be fixed by replacing ``the'' with ``any''.
}
We refer to this \defn{simple} collection of \defn{datasets} as \defn{\simpleDatasetName}. \simpleDatasetName is representative of datasets used in past work, which are not used directly as prior work does not use all datasets comprising \datasetName.

\paragraph{High-level Comparison.}\label{sec:dataset:compare}
\cref{tab:dataset-properties} summarizes different properties of related datasets.
A comparison between the volume of questions generated with other template approaches is provided in \cref{tab:dataset-comparison} for the five source datasets featured in \datasetName.
We give example questions from ID and UReader in \cref{app:examples}. In \cref{tab:dataset-comparison}, we see that \datasetName has $3$-$6$ times more templates than ID and UReader. The total number of questions and number of questions per document are also consequently higher.


\input{tables/perplexity}

\paragraph{Diversity and Realism Comparison.} To quantitatively assess how closely \datasetName resembles human-annotated datasets compared to \simpleDatasetName, we conduct two studies.
First, we measure the perplexity of the questions of the DocVQA and DUDE test sets with a small language model (GPT2) fine-tuned on the questions of \datasetName and \simpleDatasetName separately.
Perplexity indicates how likely a language model is to generate new input, so a low perplexity in this experiment suggests that the unseen human-generated questions from DocVQA and DUDE align with the distribution of the fine-tuned GPT2.
Secondly, following \citet{ye-2022-zerogen}, we compute the self-BLEU score of the questions of the \datasetName and \simpleDatasetName test sets to compare their diversity to that of DUDE and DocVQA questions. See \cref{app:metrics} for more details on the perplexity and self-BLEU experiments.

\cref{tab:perplexity} gives the perplexity scores of GPT2 fine-tuned and tested on each combination of datasets and the self-BLEU scores.
We observe that, even though DUDE and DocVQA generalize well to each other, \datasetName and \simpleDatasetName have a much higher perplexity. Similarly, the self-BLEU scores of \datasetName and \simpleDatasetName are higher than those of DUDE and DocVQA -- highlighting the challenges of mimicking human-crafted questions using templates in general.
However, fine-tuning GPT2 on \datasetName questions did reduce the perplexity from fine-tuning on \simpleDatasetName by a factor of $4$ and $7$ on DocVQA and DUDE respectively.
Additionally, the self-BLEU score of \datasetName is closer to that of DocVQA and DUDE than \simpleDatasetName.
These results demonstrate the benefits of \datasetName over simpler templating methods in terms of similarity to human-curated data.

\input{tables/main_results}

%% file: tables/dataset_properties.tex
\begin{table}[t]
    \centering
    \small
    \begin{tabular}{l cccc}
         \bf Property & \bf DUDE & \bf ID & \bf UReader & \bf \datasetName \\ \midrule
         Templated & \xmark & \cmark & \cmark & \cmark \\
         Diverse & \cmark & \cmark & \xmark & \cmark \\
         Extractive & \cmark & \xmark & \cmark & \cmark \\
         Boolean & \cmark & \xmark & \xmark & \cmark \\
         Unambiguous & \cmark & \cmark & \xmark & \cmark
    \end{tabular}
    \caption{Properties of VRDU datasets for generative models. DUDE is provided here as a human-generated reference.}
    \label{tab:dataset-properties}

    \vspace{-1em}
\end{table}

%% file: tables/validation.tex
\begin{table}[t]
    \centering
    \small
    \begin{tabular}{l rr}
         \bf Error Type  & \bf Error rate & \bf Fleiss $\boldsymbol{\kappa}$ \\ \midrule
         Template & $1.15\%$ & $0.53$ \\
         Cleaning & $1.38\%$ & $0.31$ \\
         Annotation & $2.76\%$ & $0.51$ \\
         Other & $0.23\%$ & $0.51$ \\
         \midrule
         \bf Total & $5.52\%$ & $0.49$ 
    \end{tabular}
    \caption{\datasetName validation results. We break down errors into four categories: (1) Template errors indicate a question design issue, (2) Cleaning errors indicate a data cleaning issue, (3) Annotation errors indicate issues with the original datasets, and (4) Other errors (e.g., OCR issues).}
    \label{tab:validation}

    \vspace{-1em}
\end{table}

%% file: tables/dataset_comparison.tex
\begin{table}[t]
    \centering
    \small 
    \begin{tabular}{
    L{1.2cm} L{1.3cm} C{0.8cm} C{1.1cm} C{1.2cm}
    }
        \bf Source Dataset & \bf New Dataset & \bf Num. Temp. & \bf Num. Ques. & \bf Ques. per Doc.
        \\ 
        \midrule
        \multirow{2}{1.2cm}{Ad-Buy}
        & \datasetName & $50$ & $15,\!119$ &$23.6$ \\
        & \simpleDatasetName & $1$ &  $4,\!986$ & $7.8$ \\
        \midrule
        \multirow{3}{*}{CORD}
        & \datasetName & $22$ & $39,\!575$ & $39.6$ \\
        & \simpleDatasetName & $1$ & $4,\!143$ & $4.1$ \\
        & ID & $5$ & $1,336$ & $1.3$ \\
        \midrule
        \multirow{3}{*}{Docile}
        & \datasetName & $17$ & $185,\!557$ & $32.7$ \\
        & \simpleDatasetName & $1$ & $53,\!547$ & $9.4$ \\
        & ID & $5$ & $56,\!369$ & $9.9$ \\
        \midrule
        \multirow{4}{*}{KLC}
        & \datasetName & $31$ & $44,\!813$ & $16.1$ \\
        & \simpleDatasetName & $1$ & $19,\!348$ & $7.0$ \\
        & ID & $5$ & $13,\!449$ & $4.8$ \\
        & UReader & $1$ & $27\!,664$ & $8.0$ \\
        \midrule
        \multirow{2}{1.2cm}{Reg. Form}
        & \datasetName & $18$ & $23,\!427$ & $12.2$ \\
        & \simpleDatasetName & $1$ & $8,826$ & $4.6$ \\
    \end{tabular}
    \caption{Comparison of variety of questions in \datasetName compared to \simpleDatasetName, ID, and UReader.
    Examples from each dataset are given in \cref{app:examples}.}
    \label{tab:dataset-comparison}
    \vspace{-1em}
\end{table}

%% file: tables/perplexity.tex
\begin{table}[t]
    \centering
    \small
    \begin{tabular}{l rr rr}
         \multirow{2}{*}{\bf Dataset} & \multicolumn{2}{c}{\bf Perplexity ($\downarrow$)} & \multicolumn{2}{c}{\bf Self-BLEU ($\downarrow$)} \\ \cmidrule(lr){2-3} \cmidrule(lr){4-5}
         & \bf DUDE & \bf DocVQA & \bf $\boldsymbol{2}$-gram & \bf $\boldsymbol{4}$-gram \\
         \midrule
         DUDE & $28.5$ & $35.9$ & $0.73$ & $0.40$ \\
         DocVQA & $45.1$ & $27.4$ & $0.83$ & $0.58$ \\
         \datasetName & $229.6$ & $228.6$ & $0.92$ & $0.83$ \\
         \simpleDatasetName & $1592.8$ & $928.5$ & $1.00$ & $1.00$ \\
    \end{tabular}
    \caption{Perplexity and Self-BLEU scores.
    For the perplexity experiments, GPT2 is fine-tuned for one epoch on \datasetName and on \simpleDatasetName, two on DUDE, and three on DocVQA to match the number of training steps.}
    \label{tab:perplexity}
    \vspace{-1em}
\end{table}


%% file: tables/main_results.tex
\begin{table}[t]
    \small
    \centering
    \begin{tabular}{L{1.2cm} R{0.7cm} R{.95cm} R{.85cm} R{.7cm} R{0.7cm}}
         \bf Model & \bf Ad-Buy & \bf CORD & \bf Docile & \bf KLC & \bf Reg. Form \\ \midrule
         Dnt$_{\text{ZS}}$ & $1.4$ & $18.0$ & $7.2$ & $3.8$ & $5.9$ \\
         P2S$_{\text{ZS}}$ & $13.5$ & $27.0$ & $24.1$ & $11.8$ & $24.0$ \\
         Doc$_{\text{ZS}}$ & $23.9$ & $43.0$ & $48.0$ & $\boldsymbol{80.4}$ & $27.6$ \\
      UDOP$_{\text{ZS}}$ & $29.1$ & $29.7$ & $35.1$ & $30.0$ & $39.1$ \\

         mPD$_{\text{ZS}}$ & $60.9$ & $64.5$ & $56.0$ & $66.8$ & $65.2$ \\
         mPDC$_{\text{ZS}}$ & $61.9$ & $64.4$ & $53.4$ & $66.0$ & $66.8$ \\
         GPT-4$_{\text{ZS}}$ & $\boldsymbol{72.7}$ & $\boldsymbol{85.3}$ & $\boldsymbol{61.3}$ & $68.1$ & $\boldsymbol{76.5}$ \\
         \midrule
         Dnt$_{\text{\simpleDatasetName}}$ & $7.7$ & $24.4$ & $22.3$ & $15.6$ & $19.7$ \\
         P2S$_{\text{\simpleDatasetName}}$ & $18.0$ & $31.0$ & $35.5$ & $28.0$ & $29.7$ \\         
Doc$_{\text{\simpleDatasetName}}$ & $\boldsymbol{46.3}$ & $\boldsymbol{43.6}$ & $\boldsymbol{53.0}$ & $\boldsymbol{62.8}$ & $\boldsymbol{66.2}$ \\
         \midrule
         
         Dnt$_{\text{\sampledDatasetName}}$ & $56.4$ & $82.7$ & $47.7$ & $67.2$ & $81.4$ \\
         P2S$_{\text{\sampledDatasetName}}$ & $69.2$ & $84.7$ & $59.0$ & $77.0$ & $\boldsymbol{87.0}$ \\              
    Doc$_{\text{\sampledDatasetName}}$ & $\boldsymbol{92.6}$ & $\boldsymbol{94.0}$ & $\boldsymbol{88.9}$ & $\boldsymbol{92.6}$ & $78.0$ \\
         \midrule
         Dnt$_{\text{\datasetName}}$ & $58.5$ & $83.3$ & $47.8$ & $68.1$ & $82.7$ \\
         P2S$_{\text{\datasetName}}$ & $73.8$ & $86.5$ & $59.5$ & $79.4$ & $\boldsymbol{88.7}$ \\
         Doc$_{\text{\datasetName}}$ & $\boldsymbol{93.9}$ & $\boldsymbol{96.5}$ & $\boldsymbol{90.0}$ & $\boldsymbol{93.6}$ & $80.3$ \\
    \end{tabular}
    \caption{ANLS ($\uparrow$) results on \datasetName test set using various training settings for models Donut (Dnt), Pix2Struct (P2S), DocLLM (Doc), UDOP, mPlugDocOwl 1.5 (mPD), mPlugDocOwl 1.5-Chat (mPDC), and GPT-4.
    }
    \label{tab:main-results}

    \vspace{-1em}
\end{table}

%% file: sections/4experiments.tex
\section{Modeling Experiments}\label{sec:experiments}

\input{figures/template_deltas}
\subsection{Benchmark Models}\label{sec:experiments:models}
To analyze the impact of training and testing on \datasetName, we consider three non-LLM generative models in our experiments: Donut (200M parameters) \citep{kim-2022-ocr} Pix2Struct base (282M parameters) \citep{lee-2023-pix2struct}, and UDOP (800M parameters) \citep{tang-23-unifying}, which uses OCR-generated text input as well as the image of the document.
Additionally, we consider four LLMs.
Firstly, mPlugDocOwl 1.5 (8.1B parameters) and mPlugDocOwl 1.5-chat (8.1B parameters) \citep{hu-2024-docowl}, which are OCR-free LLMs and use the image of the document as input.
Secondly, DocLLM (1.5B parameters) \citep{wang2023docllm} and the text-only variant of GPT-4\footnote{We use \texttt{gpt-4-0613} prompted with text tokens. Other models are prompted as described in their respective papers.} \citep{gpt4} , which are OCR-based.
GPT-4, UDOP, and the mPlugDocOwl models are only run in the zero-shot experiments due to resource constraints and data contamination considerations. Indeed, UDOP and the mPlugDocOwl models incorporate some of our original datasets in their pretraining; this may also be the case for GPT-4. 
We initialize all trainable models from DocVQA-fine-tuned checkpoints.
Further model set-up details are given in \cref{app:experiments}. 

\subsection{Evaluation Settings}\label{sec:experiments:metrics}

We consider several training data settings for our modeling experiments.
Firstly, we focus on the full \datasetName dataset as presented in \cref{sec:dataset}.
Secondly, a baseline dataset, \simpleDatasetName (as described in \cref{sec:dataset:simple}), is used to quantify the importance of having richer and more diverse templates.
\datasetName offers two main advantages over \simpleDatasetName: (1) the complexity of the questions and (2) the size of the dataset.
To measure the impact of each advantage, we also train the models on a down-sampled version of \datasetName, which we call \defn{mini-\datasetName} (\defn{\sampledDatasetName}).
This dataset aims to reflect the same complexities as \datasetName but with a comparable size to \simpleDatasetName.
Note that questions that span multiple pages are not used in our experiments, as not all models have multi-page capabilities.

We evaluate model performance using the Average Normalized Levenshtein Similarity (ANLS) metric \citep{Biten_2019_ICCV}. We define the ANLS score in \cref{app:metrics}.
We chose ANLS as the performance measure because it focuses on the surface similarity of the generated and true answer (as is intended in extraction) but does not suffer from the rigidity of exact matches.


\subsection{Model Performance on 
\datasetName}\label{sec:experiments:results}

\cref{tab:main-results} gives the benchmark results for \datasetName.\footnote{Results on \simpleDatasetName are given in \cref{app:results:simple} for comparison.}
As expected, training on simple templates (\simpleDatasetName) and testing on \datasetName questions yields much lower performance than training directly on \datasetName or \sampledDatasetName.
We delve into the important effects of training and testing on different datasets in \cref{sec:experiments:deltas}. 
In general, models trained on simple templates perform slightly better than their zero-shot counterparts, except DocLLM trained on KLC with simple templates and tested on K2Q templates.
We also observe a consistently positive impact of training on \datasetName versus its down-sampled counterpart \sampledDatasetName across the board (1.5\% for Donut, 3\% for Pix2Struct, and 1.9\% for DocLLM on average). This highlights the diversity of questions contained in \datasetName as well as the impact of greater training volume enabled by a template-based approach.
These trends support our hypothesized benefits of \datasetName.
\cref{app:results:ablation} contains an ablation study on how the different training configurations impact performance on extractive and boolean questions and the number of entities present in each question.

We typically observe performance increases with model size on the same data setting.
Indeed, DocLLM performs best among the models trained, followed by Pix2Struct and Donut.
In the zero-shot setting, GPT-4 is the best model on all datasets except KLC. 
There is a large gap between the zero-shot and the fine-tuned results for Pix2Struct, Donut, and DocLLM -- despite the original checkpoints we used being fine-tuned on DocVQA.
This indicates sensitivity to distribution shifts in the type of documents and questions.



\subsection{Post-training Assessment of Model Robustness to Template Change}\label{sec:experiments:deltas}

\input{figures/groundedness/grounded_klc}
This work hypothesizes that using the same simple template for all datasets for training or evaluation does not reflect the intricacies of document understanding and may reduce model robustness at test time.
To address this, we propose a suite of rich and diverse templates for transforming each dataset to create \datasetName.
To validate this hypothesis and motivate the need for our work, \textit{we compare how models trained on simple templates perform when tested on \datasetName templates, and vice versa}.
To evaluate this experiment, we define the following metric that quantifies model robustness at test time to change in the templating approach used at train time:
\begin{equation}
    \Delta_{D_1, D_2} \defeq \frac{\mathrm{ANLS}_{D_1/D_1} - \mathrm{ANLS}_{D_2/D_1}}{\mathrm{ANLS}_{D_1/D_1}}
\end{equation}
where $D_1$ and $D_2$ are datasets drawn from different distributions, and $\mathrm{ANLS}_{D_2/D_1}$ is the ANLS score of a model trained on the train split of $D_2$ and evaluated on the test split of $D_1$.
$\mathrm{ANLS}_{D_1/D_1} \geq \mathrm{ANLS}_{D_2/D_1}$ most often, as training and testing on datasets drawn from the same distribution almost always leads to better performance. Thus, $\Delta_{D_1, D_2}$ measures the change in performance when swapping $D_1$ and $D_2$ for training while testing on $D_1$. A large value of $\Delta_{D_1, D_2}$ indicates that training on $D_2$ cannot generalize well to $D_1$. Conversely, a low value indicates that training on $D_2$ or $D_1$ generalizes well to $D_1$.

\cref{fig:template-deltas} shows the difference between $\Delta_{\text{\sampledDatasetName}, \text{\simpleDatasetName}}$ and $\Delta_{\text{\simpleDatasetName}, \text{\sampledDatasetName}}$ for the three trainable models.
We use \sampledDatasetName (the down-sampled version of \datasetName introduced in \cref{sec:experiments:metrics}) 
to enable a fair comparison based on template style and not affected by training data volume.
We observe in \cref{fig:template-deltas} that the red bars corresponding to $\Delta_{\text{\simpleDatasetName}, \text{\sampledDatasetName}}$ are lower than the blue bars corresponding to $\Delta_{\text{\sampledDatasetName}, \text{\simpleDatasetName}}$ in 14 out of 15 cases.
Consequently, training on \sampledDatasetName tends to yield better generalization to \simpleDatasetName than the other way around.
Indeed, a difference of $45\%$ is observed when comparing $\Delta_{\text{\sampledDatasetName}, \text{\simpleDatasetName}}$ to $\Delta_{\text{\simpleDatasetName}, \text{\sampledDatasetName}}$ on average across all datasets and models (see green and red arrows in \cref{fig:template-deltas}).
This observation corroborates the motivation behind \datasetName that employing a rich and diverse set of templates for training models results in better robustness to new types of questions and formulations.
We provide more metrics, including a comparison with the full \datasetName, in \cref{app:results:templates}.

\subsection{Evaluation of Generated Errors}\label{sec:experiments:grounding}

While ANLS is a useful metric for comparing extractive results from generative models, it fails to determine the cause of the errors, such as misreading the text in the document or not understanding the question.
We use the notion of \defn{groundedness} to determine this breakdown.
A generated response is considered grounded if it can be identified in the OCR output of the document.\footnote{We consider a response to be in the OCR if we can find a non-case-sensitive match.
The match can be found within an OCR token or across OCR tokens.}
Correct generations are naturally grounded.
We define incorrect generations that are grounded as \defn{mis-extractions}.
For ungrounded generations, we consider extracted strings that have an ANLS score of $0.8$ or higher to be \defn{misprints}.
Any other errors we label as \defn{other}. 
We provide examples in \cref{app:results:error_examples}.

\cref{fig:grounded:klc} gives the breakdown of groundedness and error types for the KLC extractive questions of \datasetName using Donut, Pix2Struct, and DocLLM finetuned on \datasetName.
We choose KLC due to its large test set.
The models tested on \datasetName typically exhibit a higher level of grounding than those tested on \simpleDatasetName, regardless of the training data.
This could suggest that formulating templates for the dataset rather than using generic templates allows the models to contextualize the questions better. Prompting the model with questions framed using entities present in the document likely guides the model in generating answers that are also present in the document. Grounded responses enable easier verification of KIE model outputs.
Note that both Donut and Pix2Struct contain a large number of misprint errors compared to DocLLM.
This could mean that using a more powerful OCR-free method may result in much stronger performance.

%% file: figures/template_deltas.tex
\begin{figure*}[t]
    \centering     \includegraphics[width=\textwidth]{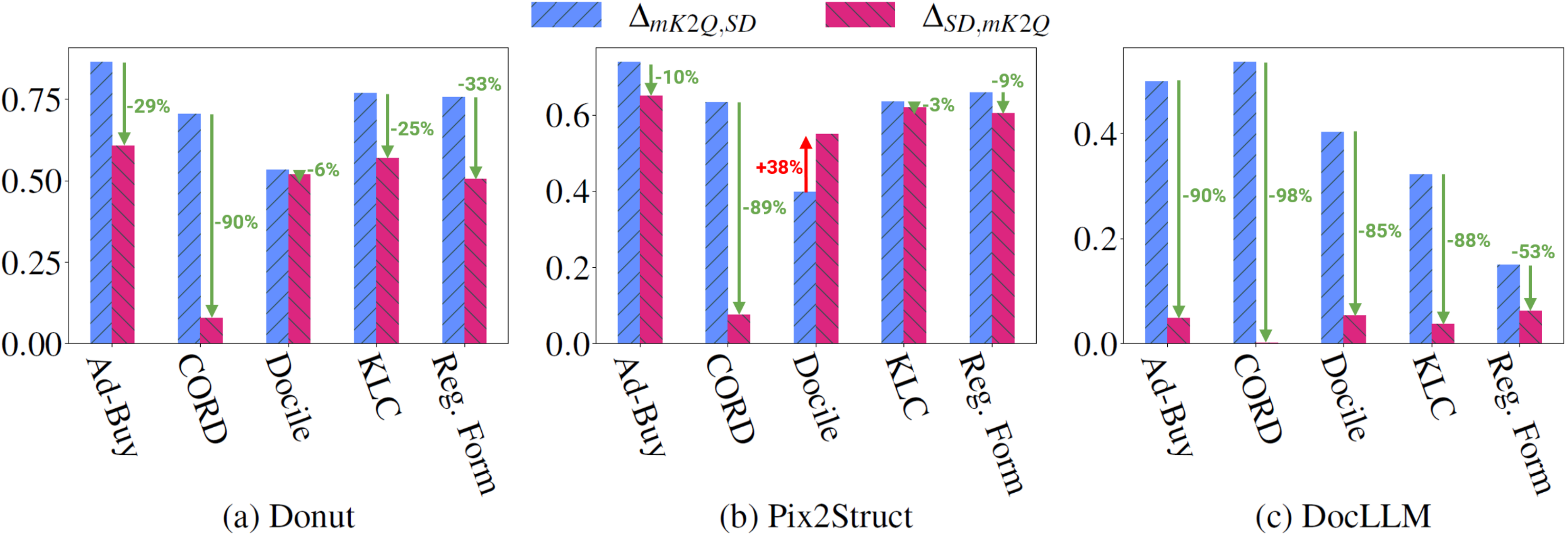}
    \vspace{-1.5em}
    \caption{Comparison of training and evaluating on complex questions (\sampledDatasetName) and simple questions (\simpleDatasetName). }
    \label{fig:template-deltas}

    \vspace{-1em}
\end{figure*}

%% file: figures/groundedness/grounded_klc.tex
\begin{figure*}[t]
    \centering     \includegraphics[width=\textwidth]{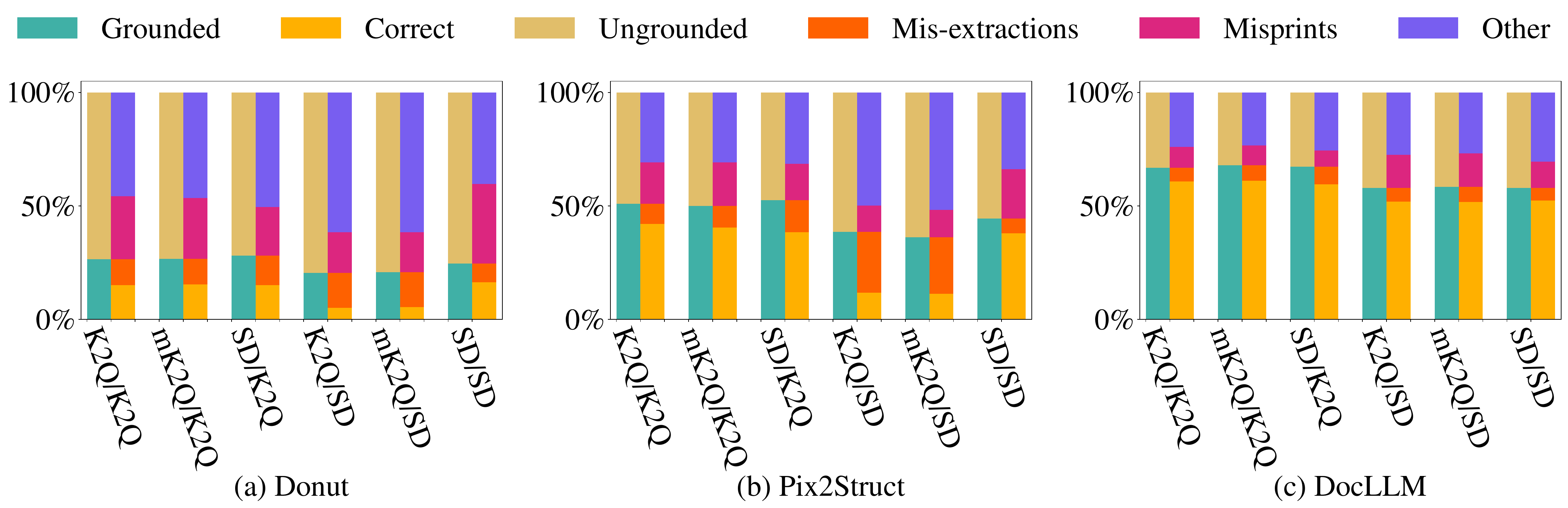}
    \vspace{-1em}
    \caption{Detailed breakdown of groundedness and error types for KLC using different training/testing datasets.}
    \label{fig:grounded:klc}

    \vspace{-1em}
\end{figure*}

%% file: sections/5conclusions.tex
\section{Conclusion}\label{sec:conc}

This paper introduced \datasetName, a new publicly available collection of five transformed KIE datasets for generative VRDU models.
\datasetName provides a large and diverse set of template-based questions that better capture the intricacies of KIE and the variety of questions that users can ask in real-world applications. We present a middle ground between LLM-generated and manually-curated questions for instruction tuning, which trades off the time to craft such data against diversity. Our approach can be extended to domains such as general VQA, multi-turn multimodal conversation, and video QA, which we leave to future work.
Our experiments demonstrate that training generative models on \datasetName instead of data from simple templates improves generalization to held-out types of instructions.
In addition, our error analysis suggests that questions in \datasetName provide enhanced contextualization compared to simple ones, resulting in more grounded answers from models, regardless of correctness.
In future work, we plan to create templates covering a wider range of question types, few-shot instances, chain-of-thought answers with layout-informed explanations, and multi-round instructions.
We hope this work encourages researchers to carefully consider 
the data used for generative modeling.


%% file: sections/limitations.tex
\section*{Limitations}


We conducted our experiments using three trainable models (Donut, Pix2Struct, and DocLLM) and four zero-shot models (UDOP, mPlugDocOwl 1.5, mPlugDocOwl 1.5-chat, and GPT-4).
With additional resources, training state-of-the-art OCR-free models such as the mPlugDocOwl models would provide more complete results.
Mainly due to resource limitations for training vision-based LLMs, we left these experiments (along with training UDOP)
for future work. We also note that mPlugDocOwl, UDOP, and possibly GPT-4 have already seen some of the datasets used in this work during the pretraining phase. Thus, data contamination could affect the zero-shot and fine-tuning performance of these models.

Additionally, while \datasetName alleviates the burden of data collection by relying on existing KIE datasets, it still requires human intervention to curate high-quality, diverse templates manually. 
We will investigate using LLMs such as GPT-4 to generate templates for VRDU applications in future work.  

%% file: sections/acknowledgements.tex
\section*{Disclaimer}
This paper was prepared for information purposes by the Artificial Intelligence Research group of JPMorgan Chase \& Co and its affiliates (“JP Morgan”) and is not a product of the Research Department of JP Morgan. J.P. Morgan makes no representation and warranty whatsoever and disclaims all liability for the completeness, accuracy, or reliability of the information contained herein. This document is not intended as investment research or investment advice, or a recommendation, offer, or solicitation for the purchase or sale of any security, financial instrument, financial product, or service, or to be used in any way for evaluating the merits of participating in any transaction, and shall not constitute a solicitation under any jurisdiction or to any person if such solicitation under such jurisdiction or to such person would be unlawful.

%% file: appendices/a2methodology.tex
\section{Dataset Construction}\label{app:dataset}

\subsection{Data Cleaning}\label{app:dataset:clean}
To make questions and answers sound natural, we apply several data cleaning rules to the KIE annotations in the datasets instead of using the OCR text.
We describe the data cleaning method for each dataset below.

\paragraph{Ad-Buy.}
For all entities, we replace all new line characters with spaces except for addresses where we insert a comma.
Next, we check if the text is enclosed in brackets (parentheses, square, braces, etc.) and if so, we remove the brackets.
In addition, we found that some entities were prefixed with ``REMIT TO'', so we deleted any occurrence of this prefix.
For the \texttt{advertiser}, \texttt{agency}, \texttt{product}, \texttt{program\_desc}, and \texttt{tv\_address} entities, we convert all multi-word, full upper case entities into title case.
We have an additional caveat for \texttt{tv\_address} and do not modify words less than two characters as they are likely US states.

\paragraph{CORD.}
CORD contains too many entities to opt for an entity-type approach.
However, two-thirds of all entities are numeric.
For these entities, we remove all OCR tokens for which over half the characters are non-numeric.
We also remove any leading or trailing punctuation symbols from the entities.

\paragraph{Docile.}
We take a similar approach to cleaning Docile entities as with CORD due to the similarities in their annotation schemes.
For numeric entities, we consider only numerical and punctuation symbols.
We treat numbers differently in Docile than in CORD due to our dataset-specific analyses of the original annotations.
For non-numeric entities, we replace newline characters with a comma and a space.
Additionally, we pre-process the text to the title case.
We acknowledge that this may not always be the correct choice of data cleaning.

\paragraph{KLC.}
For KLC, we only title case entities that are  \texttt{address\_post\_town}, \texttt{address\_street\_line}, and \texttt{charity\_name}.
The other entities are identification type entities (e.g., \texttt{charity\_name}) or numbers and so data cleaning may compromise their values.

\paragraph{Reg. Form.}
We apply data cleaning to Reg. Form using the same methods described for Ad-Buy. In addition, we apply additional checks for words that should not be converted to title case. For each entity type, we find common acronyms used in the text (e.g., USA, LLC) and ensure these tokens remain in all caps.


\subsection{Data Splits}\label{app:dataset:split}
We follow the data splits provided by the original datasets.
For Docile, the downloadable test set does not give the annotated entities; we thus use the dev set as its test set.
For Ad-Buy and Reg. Form, multiple splits were given in \citet{wang-2023-vrdu}, so we follow the split in \citet{wang2023docllm}.
We provide the number of questions and documents for each dataset split in \cref{tab:dataset-split}.

\input{tables/dataset_split}

\subsection{Error type definitions in validation exercise}\label{app:dataset:errors}

\cref{tab:validation} reports the various types of errors we found in the template-generated questions. We define these errors in more detail below, along with guidelines provided to annotators.


\begin{enumerate}
    \item Template Error: This is a grammatical mistake in the template. If the question doesn’t make sense after plugging in the specific values, then it is marked as a template error.
    \item Cleaning Error: This is a mistake in post-processing the value, such as changing the case. The question would make sense if the original value from the document was used.
    \item Annotation Error: This is a mistake in the original dataset where an entity should have been annotated differently. For example, “What was the address of the company named 123?” where 123 is an ID but was tagged as a name.
    \item Other Error: The question doesn’t make sense for a reason not included in the above errors.

\end{enumerate}

%% file: tables/dataset_split.tex
\begin{table}[t]
    \centering
    \small
    \begin{tabular}{l ccc}
         \bf Dataset & \bf Train & \bf Dev & \bf Test \\ \midrule
         \multirow{2}{*}{Ad-Buy} & $11,\!362$ & $-$ & $3,\!757$ \\
          & ($480$) & ($-$) & ($161$) \\ \midrule 
         \multirow{2}{*}{CORD} & $62,\!948$ & $7,\!242$ & $7,\!551$ \\
          & ($800$) & ($100$) & ($100$) \\ \midrule
         \multirow{2}{*}{Docile} & $169,\!664$ & $-$ & $15,\!893$ \\
          & ($5,\!180$) & ($-$) & ($500$) \\ \midrule
          \multirow{2}{*}{KLC} & $27,\!180$ & $7,\!294$ & $10,\!339$ \\
          & ($1,\!729$) & ($440$) & ($609$) \\ \midrule
          \multirow{2}{*}{Reg. Form} & $17,\!539$ & $-$ & $5,\!888$ \\
          & ($1,\!436$) & ($-$) & ($479$) \\ \midrule
    \end{tabular}
    \caption{Number of questions and documents (in brackets) for \datasetName splits.}
    \label{tab:dataset-split}
\end{table}

%% file: appendices/a1templateexamples.tex
\section{\datasetName Examples}\label{app:examples}

In this section, we provide example documents and questions generated from each of the five datasets of \datasetName.
\cref{fig:examples:adbuy}, \cref{fig:examples:cord}, \cref{fig:examples:docile}, \cref{fig:examples:klc},  and \cref{fig:examples:regform} compare questions generated in \datasetName to those found in InstructDoc \citep{tanaka-2024-instructdoc}, UReader \citep{ye-2023-ureader}, and \simpleDatasetName.
Note that InstructDoc and UReader do not contain most of the datasets used in this work.
We thus use their templates to generate questions as described in the respective papers.

\input{figures/appendix_examples/ad_buy_example}
\input{figures/appendix_examples/cord_example}
\input{figures/appendix_examples/docile_example}
\input{figures/appendix_examples/klc_example}
\input{figures/appendix_examples/reg_form_example}
The key differences between the questions found in \datasetName and those found in UReader and \simpleDatasetName are the style and informativeness of the questions.
Our templates are designed with the dataset in mind and thus sound more natural and resemble questions that one might encounter in the real world.
Furthermore, our templates use other key entities within the questions to better contextualize queries as well as enable a wider selection of questions.
We note that for datasets such as Docile which contain line items, UReader may generate ambiguous questions.
For example, the question regarding the line item total price in \cref{fig:examples:docile} has a variety of answers that could be correct as it is unclear which line was being referred to.
Docile contains documents that have many lines, so such under-specified questions are avoided.

The primary difference between \datasetName and InstructDoc lies in the kind of questions being asked.
As explained in \cref{sec:dataset:compare}, InstructDoc transforms the information extraction problem into a classification problem.
Therefore, the KIE portion of InstructDoc trains and tests a model's ability to characterize a key entity to its correct type, but not to find the entity itself.
Other than formulating questions for a different task, InstructDoc has five dataset-agnostic templates it populates.
Consequently, the variety of questions in \datasetName is much wider.

%% file: figures/appendix_examples/ad_buy_example.tex
\definecolor{color1}{HTML}{648FFF}
\definecolor{color2}{HTML}{785EF0}
\definecolor{color3}{HTML}{DC267F}
\definecolor{color4}{HTML}{FE6100}
\definecolor{color5}{HTML}{FFB000}
\definecolor{color6}{HTML}{117733}

\begin{figure*}[]
\centering
\begin{tikzpicture}

\node[draw=gray, rounded corners, minimum width=15.5cm, minimum height=11.5cm] at (0, -2.2) {};

\node[draw=none,fill=none] at (0,0.1){\includegraphics[width=0.95\textwidth]{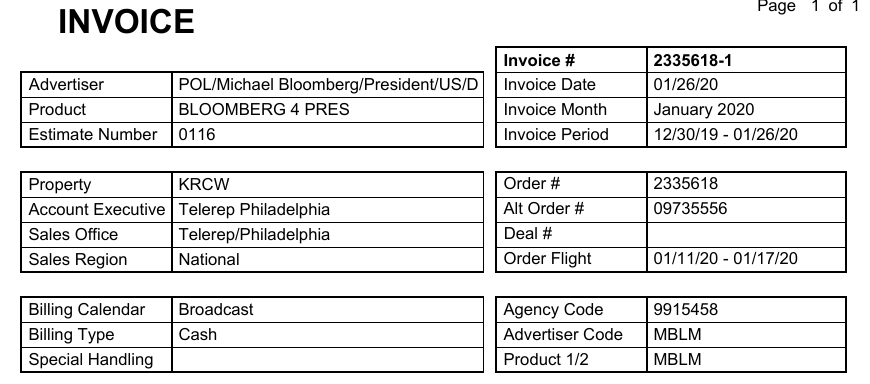}};

\node[draw=color1, rounded corners, minimum width=1.3cm, minimum height=0.35cm, line width=0.4mm] at (4.38, -1.07) {};
\node[draw=color1, rounded corners, minimum width=1.3cm, minimum height=0.35cm, line width=0.4mm] at (5.78, -1.07) {};
\node[draw=color2, rounded corners, minimum width=3.2cm, minimum height=0.38cm, line width=0.4mm] at (-2.95, 1.52) {};
\node[draw=color6, rounded corners, minimum width=5.35cm, minimum height=0.36cm, line width=0.4mm] at (-1.88, 1.95) {};

\node[draw=color3, rounded corners, minimum width=9.3cm, minimum height=2.2cm] at (-2.8, -4.4) {};
\node[] at (-5.9, -3.75) {\bf \datasetName Questions:};
\node[align=left] at (-4.1, -4.37) {\small \textsc{Q}: From when until when is the contract in {\color{color1}flight}? \\[-.4em] \small \textsc{A}: 01/11/20 - 01/17/20};
\node[align=left] at (-3.54, -5.15) {\small \textsc{Q}: Did {\color{color6}POL/Michael Bloomberg/President/US/D} order the \\[-.4em] \small advertisement {\color{color2}``Bloomberg 4 Pres''}? \textsc{A}: Yes};

\node[draw=color4, rounded corners, minimum width=9.3cm, minimum height=1.9cm] at (-2.8, -6.6) {};
\node[] at (-4.7, -6.) {\bf UReader and \simpleDatasetName Questions:};
\node[] at (-3.14, -6.6) {\small \textsc{Q}: What is the value for the {\color{color6} product}? \textsc{A}:  BLOOMBERG 4 PRES};
\node[] at (-3.63, -7.1) {\small \textsc{Q}: What is the value for the {\color{color1} flight start date}? \textsc{A}: 01/11/20};

\node[draw=color5, rounded corners, minimum width=5.3cm, minimum height=4.3cm] at (4.8, -5.4) {};
\node[] at (4.3, -3.8) {\bf InstructDoc Questions:};
\node[align=left] at (4.4, -5.5) {\small \textsc{Q}: There are 15 categories for \\[-.3em] \small selection: ``advertiser'', \dots, and  \\[-.3em] \small ``tv\_address''. Please output the \\[-.3em] \small category corresponding to the \\[-.3em] \small text {\color{color6}``BLOOMBERG 4 PRES''}.  \\ \small \textsc{A}: product};

\end{tikzpicture}
\caption{Excerpt of an Ad-Buy document with generated questions from \datasetName, InstructDoc, UReader, and \simpleDatasetName. The K2Q question ``From when until when is the contract in flight?'' uses jargon specific to the advertising domain. Applying such templates allows for creating domain-specific and diverse questions, which may differ from what is colloquially used. The generated question is thus grounded in the jargon used in the document.}
\label{fig:examples:adbuy}

\end{figure*}

%% file: figures/appendix_examples/cord_example.tex
\definecolor{color1}{HTML}{648FFF}
\definecolor{color2}{HTML}{785EF0}
\definecolor{color3}{HTML}{DC267F}
\definecolor{color4}{HTML}{FE6100}
\definecolor{color5}{HTML}{FFB000}
\definecolor{color6}{HTML}{117733}
\definecolor{color7}{HTML}{AA4499}

\begin{figure*}[]
\centering
\begin{tikzpicture}

\node[draw=none,fill=none] at (0,0.1){\includegraphics[width=0.4\textwidth]{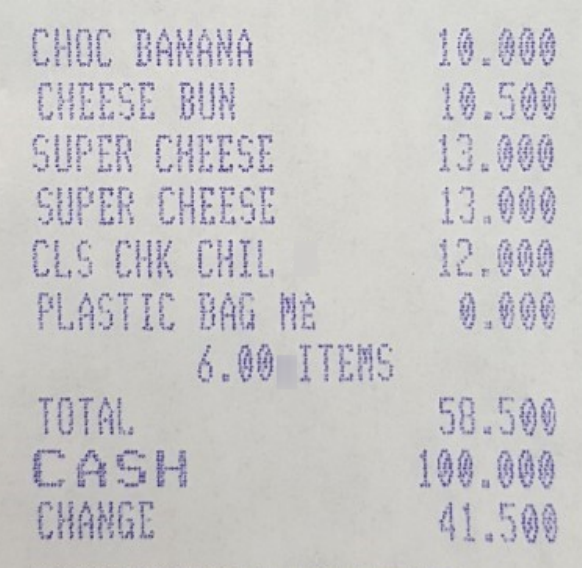}};

\node[draw=gray, rounded corners, minimum width=15.8cm, minimum height=7cm] at (4.5, .1) {};

\node[draw=color1, rounded corners, minimum width=2.7cm, minimum height=0.6cm, line width=0.4mm] at (-1.61, 2.75) {};
\node[draw=color2, rounded corners, minimum width=1.7cm, minimum height=0.6cm, line width=0.4mm] at (2.3, 2.75) {};
\node[draw=color6, rounded corners, minimum width=1.7cm, minimum height=0.6cm, line width=0.4mm] at (2.3, -2.5) {};

\node[draw=color3, rounded corners, minimum width=8.7cm, minimum height=2.7cm] at (7.8, 1.8) {};
\node[] at (4.95, 2.8) {\bf \datasetName Questions:};
\node[align=left] at (7.65, 2.2) {\small \textsc{Q}: What is the {\color{color1}menu item} that cost a total of {\color{color2}10.000} in this bill? \\[-.3em] \small \textsc{A}: CHOC BANANA};
\node[align=left] at (7.7, 1.4) {\small \textsc{Q}: How much did the order(s) of {\color{color1}CHOC BANANA} {\color{color2}cost} in total? \\[-.3em] \small \textsc{A}: 10.000};
\node[align=left] at (7.83, 0.75) {\small \textsc{Q}: Is {\color{color6}``58.500''} the amount of change in cash in this receipt? \textsc{A}: No};

\node[draw=color4, rounded corners, minimum width=8.7cm, minimum height=1.5cm] at (7.8, -0.4) {};
\node[] at (6.15, .) {\bf UReader and \simpleDatasetName Questions:};
\node[align=left] at (7.77, -.45) {\small \textsc{Q}: What is the value for the {\color{color6}amount of change in cash}? \textsc{A}: 41.500};
\node[align=left] at (7.52, -.85) {\small \textsc{Q}: What is the value for the {\color{color1}menu item}? \textsc{A}: CHOC BANANA};

\node[draw=color5, rounded corners, minimum width=8.7cm, minimum height=1.75cm] at (7.8, -2.15) {};
\node[] at (5.55, -1.55) {\bf InstructDoc Questions:};
\node[align=left] at (7.77, -2.4) {\small \textsc{Q}: Please tell me the category of the text ``41.500'' to select \\[-.2em] \small from following classes:  ``menu.nm'', \dots, and ``total.menuqty\_cnt''. \\[-.2em] \small \textsc{A}: total.changeprice};

\end{tikzpicture}
\caption{Excerpt of a CORD document with generated questions from \datasetName, InstructDoc, UReader, and \simpleDatasetName. Note that the second question for UReader and \simpleDatasetName is not present in \simpleDatasetName due to its ambiguity.}
\label{fig:examples:cord}

\end{figure*}

%% file: figures/appendix_examples/docile_example.tex
\definecolor{color1}{HTML}{648FFF}
\definecolor{color2}{HTML}{785EF0}
\definecolor{color3}{HTML}{DC267F}
\definecolor{color4}{HTML}{FE6100}
\definecolor{color5}{HTML}{FFB000}

\begin{figure*}[]
\centering
\begin{tikzpicture}

\node[draw=gray, rounded corners, minimum width=15.5cm, minimum height=13cm] at (0, -2.2) {};

\node[draw=none,fill=none] at (0,0.1){\includegraphics[width=0.95\textwidth]{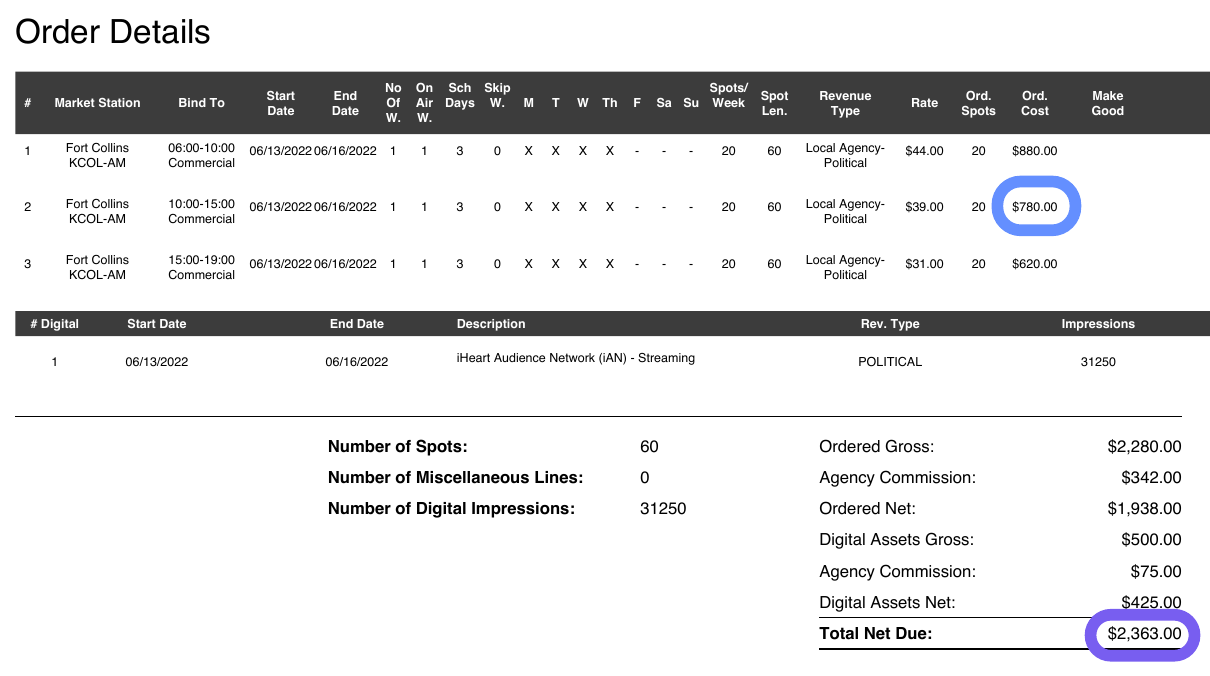}};

\node[draw=color1, rounded corners, thick, minimum width=1.3cm, minimum height=0.4cm, line width=.5mm] at (6.8, -3.6) {};
\node[draw=color2, rounded corners, thick, minimum width=1.cm, minimum height=0.35cm, line width=.5mm] at (5.4, 1.72) {};

\node[draw=color3, rounded corners, minimum width=9.3cm, minimum height=2.2cm] at (-2.8, -5.15) {};
\node[] at (-5.9, -4.55) {\bf \datasetName Questions:};
\node[] at (-2.8, -5.15) {\small \textsc{Q}: How much is the {\color{color2} total amount with tax} of the 2nd item? \textsc{A}:  780.00};
\node[] at (-2.87, -5.55) {\small \textsc{Q}: What is the {\color{color1} total amount to be paid} in the document? \textsc{A}: 2,363.00};
\node[] at (-3.05, -5.95) {\small \textsc{Q}: Is {\color{color1}``60.48''} the total amount to be paid in this document? \textsc{A}: No};

\node[draw=color4, rounded corners, minimum width=9.3cm, minimum height=1.9cm] at (-2.8, -7.4) {};
\node[] at (-4.7, -6.8) {\bf UReader and \simpleDatasetName Questions:};
\node[] at (-3.33, -7.4) {\small \textsc{Q}: What is the value for the {\color{color2} total amount with tax}? \textsc{A}:  780.00};
\node[] at (-3.1, -7.8) {\small \textsc{Q}: What is the value for the {\color{color1} total amount to be paid}? \textsc{A}: 2,363.00};

\node[draw=color5, rounded corners, minimum width=5.3cm, minimum height=4.3cm] at (4.8, -6.2) {};
\node[] at (4.4, -4.6) {\bf InstructDoc Questions:};
\node[align=left] at (4.9, -6.5) {\small \textsc{Q}: The document contains 36 key \\ \small categories: ``account\_num'', \dots, and \\ \small``total''. Kindly identify the category \\ \small related to the text {\color{color1}``2,363.00''} \\ \small mentioned in the provided document. \\ \small \textsc{A}: amount\_due};

\end{tikzpicture}
\caption{Excerpt of a Docile document with generated questions from \datasetName, InstructDoc, UReader, and \simpleDatasetName. Note that the first question for UReader and \simpleDatasetName is not present in \simpleDatasetName due to its ambiguity.}
\label{fig:examples:docile}

\end{figure*}

%% file: figures/appendix_examples/klc_example.tex
\definecolor{color1}{HTML}{648FFF}
\definecolor{color2}{HTML}{785EF0}
\definecolor{color3}{HTML}{DC267F}
\definecolor{color4}{HTML}{FE6100}
\definecolor{color5}{HTML}{FFB000}
\definecolor{color6}{HTML}{117733}
\definecolor{color7}{HTML}{AA4499}

\begin{figure*}[]
\centering
\resizebox{0.97\textwidth}{!}{
\begin{tikzpicture}

\node[draw=none,fill=none] at (0,0.1){\includegraphics[width=0.95\textwidth]{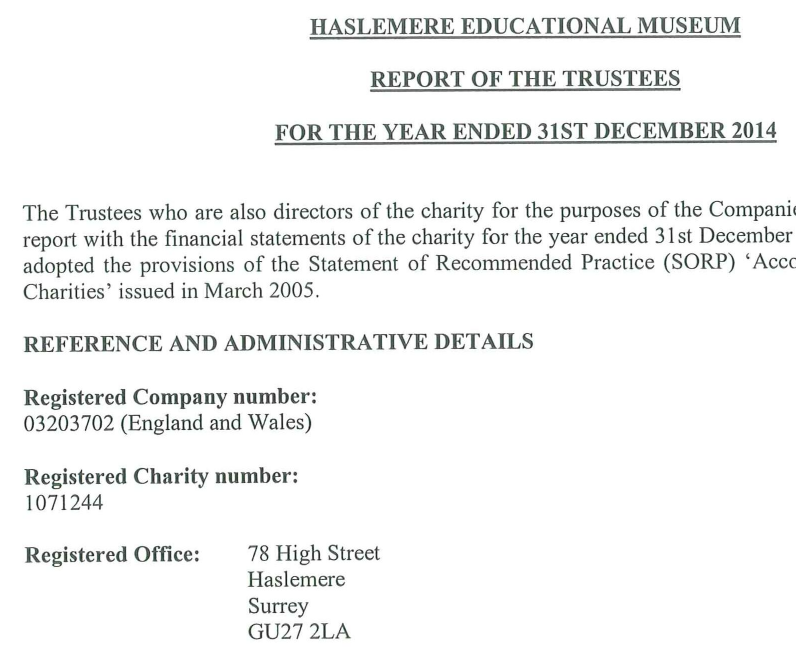}};

\node[draw=gray, rounded corners, minimum width=16cm, minimum height=14.8cm] at (0.5, -.85) {};

\node[draw=color1, rounded corners, minimum width=8.6cm, minimum height=0.7cm, line width=0.4mm] at (2.36, 5.95) {};
\node[draw=color2, rounded corners, minimum width=1.7cm, minimum height=0.44cm, line width=0.4mm] at (-6.35, -3.14) {};
\node[draw=color6, rounded corners, minimum width=2.75cm, minimum height=0.55cm, line width=0.4mm] at (-1.58, -4.08) {};
\node[draw=color7, rounded corners, minimum width=2.25cm, minimum height=0.55cm, line width=0.4mm] at (-1.8, -5.6) {};

\node[draw=color3, rounded corners, minimum width=8.3cm, minimum height=3.1cm] at (4.1, -2.15) {};
\node[] at (1.45, -.95) {\bf \datasetName Questions:};
\node[align=left] at (3.95, -1.6) {\small \textsc{Q}: What is the {\color{color2}number} of the charity {\color{color1}Haslemere Educational} \\[-.3em] \small {\color{color1} Museum}? \textsc{A}: 1071244};
\node[align=left] at (4.05, -2.4) {\small \textsc{Q}: What is the {\color{color7}postcode} for {\color{color1}Haslemere Educational Museum}? \\[-0.3em] \small \textsc{A}: GU27 2LA };
\node[align=left] at (4.15, -3.05) {\small \textsc{Q}: Is {\color{color7}GU27 2LA} the postcode of street {\color{color6}78 High Street}? \textsc{A}: Yes };

\node[draw=color4, rounded corners, minimum width=8.3cm, minimum height=2.1cm] at (4.1, -5.1) {};
\node[] at (2.65, -4.4) {\bf UReader and \simpleDatasetName Questions:};
\node[align=left] at (3.76, -4.85) {\small \textsc{Q}: What is the value for the {\color{color2}charity number}? \textsc{A}: 1071244};
\node[align=left] at (4.03, -5.3) {\small \textsc{Q}: What is the value for the {\color{color7}address postcode}? \textsc{A}: GU27 2LA};

\node[draw=color5, rounded corners, minimum width=15.6cm, minimum height=1.5cm] at (0.45, -7.3) {};
\node[] at (-5.2, -6.9) {\bf InstructDoc Questions:};
\node[align=left] at (0.05, -7.5) {\small \textsc{Q}: Categories: ``address\_postcode'', \dots, and  ``spending\_annually\_in\_british\_pounds''. Kindly provide me with the \\[-.3em] \small category of the text  {\color{color6}``78 High Street''}.  \textsc{A}: address\_street\_line};

\end{tikzpicture}
}
\caption{Excerpt of a KLC document with generated questions from \datasetName, InstructDoc, UReader, and \simpleDatasetName.}
\label{fig:examples:klc}

\end{figure*}

%% file: figures/appendix_examples/reg_form_example.tex
\definecolor{color1}{HTML}{648FFF}
\definecolor{color2}{HTML}{785EF0}
\definecolor{color3}{HTML}{DC267F}
\definecolor{color4}{HTML}{FE6100}
\definecolor{color5}{HTML}{FFB000}

\begin{figure*}[]
\centering
\resizebox{0.98\textwidth}{!}{
\begin{tikzpicture}

\node[draw=gray, rounded corners, minimum width=15.5cm, minimum height=7.5cm] at (0, -1.8) {};

\node[draw=none,fill=none] at (0,0.1){\includegraphics[width=0.95\textwidth]{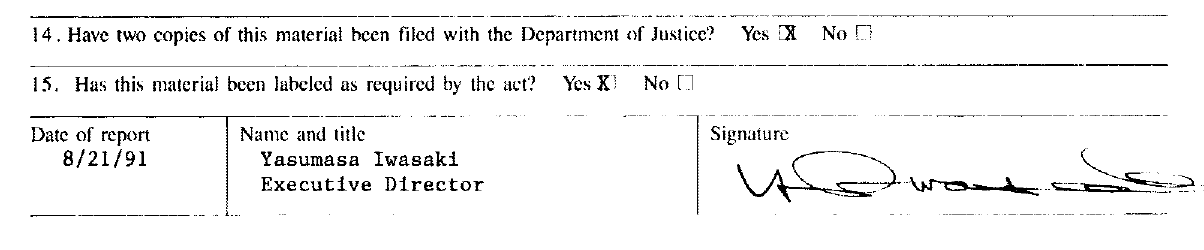}};

\node[draw=color1, rounded corners, minimum width=3.cm, minimum height=0.3cm, line width=0.4mm] at (-2.9, -0.67) {};
\node[draw=color2, rounded corners, minimum width=3.cm, minimum height=0.35cm, line width=0.4mm] at (-2.9, -0.3) {};

\node[draw=color3, rounded corners, minimum width=9.3cm, minimum height=2.2cm] at (-2.8, -2.2) {};
\node[] at (-5.9, -1.55) {\bf \datasetName Questions:};
\node[] at (-2.99, -2.15) {\small \textsc{Q}: What {\color{color2}position} does the form signer hold? \textsc{A}: Executive Director};
\node[] at (-4.17, -2.55) {\small \textsc{Q}: Did {\color{color1}Yasumasa Iwasaki} sign the form? \textsc{A}: Yes};

\node[draw=color4, rounded corners, minimum width=9.3cm, minimum height=1.9cm] at (-2.8, -4.4) {};
\node[] at (-4.7, -3.8) {\bf UReader and \simpleDatasetName Questions:};
\node[] at (-3.28, -4.4) {\small \textsc{Q}: What is the value for the {\color{color2} signer title}? \textsc{A}:  Executive Director};
\node[] at (-3.2, -4.8) {\small \textsc{Q}: What is the value for the {\color{color1} signer name}? \textsc{A}: Yasumasa Iwasaki};

\node[draw=color5, rounded corners, minimum width=5.3cm, minimum height=4.3cm] at (4.8, -3.2) {};
\node[] at (4.4, -1.6) {\bf InstructDoc Questions:};
\node[align=left] at (4.5, -3.4) {\small \textsc{Q}: Options: ``file\_date'', \dots, and \\ \small``signer\_tital''. Please select the \\ \small category associated with the text \\ \small  {\color{color1}``Yasumasa Iwasaki''} in the given \\ \small document. \\ \small \textsc{A}: amount\_due};

\end{tikzpicture}
}
\caption{Excerpt of a Reg. Form document with generated questions from \datasetName, InstructDoc, UReader, and \simpleDatasetName.}
\label{fig:examples:regform}

\end{figure*}

%% file: appendices/a3modeldetails.tex
\section{Experimental Set-up}\label{app:experiment-setup}

\subsection{Model Training Details}\label{app:experiments}
Each trainable model (Donut, Pix2Struct, and DocLLM) is then fine-tuned for ten epochs on each dataset of this study individually, with a learning rate of $10^{-4}$.
Donut and Pix2Struct use the AdaFactor optimizer with a cosine scheduler with $500$ warm-up steps and batch sizes of four and one, respectively. A weight decay of $10^{-5}$ is used.
DocLLM uses the AdamW optimizer with a cosine scheduler with warm-up and a batch size of $24$ through gradient accumulation.
We use available implementations of these models 
\citep{wolf2019huggingface} in PyTorch 1.13 \citep{paszke2017automatic}.
Each model is fine-tuned on a single A10G GPU.

\subsection{Metric Definitions}\label{app:metrics}
In this section, we provide more details regarding the perplexity, Self-BLEU, and ANLS scores used in \cref{sec:dataset:compare} and \cref{sec:experiments} respectively.

\paragraph{Perplexity.}
Perplexity measures how likely a language model is to generate a given sequence of text.
Mathematically, it is the inverse of the joint probability of a sequence of text being drawn from a distribution.
Alternatively, it is the exponentiated average of the negative log-likelihood of the input probability conditioned on all previous tokens.
If we consider a tokenized string $s=\{s_1,\dots, s_{n}\}$, the perplexity of a model $M$ is defined as
\begin{equation*}
    \mathrm{PPL}(s) \defeq \mathrm{exp}\!\left(- \frac{1}{n} \sum_{i=1}^{n} \log p_{M}(s_i\mid s_{<i})\right)
\end{equation*}
where $p_{M}(s_i\mid s_{<i})$ is the probability of witnessing token $s_i$ in the model, after observing all past tokens of $s$.

\paragraph{Self-BLEU.}
Self-BLEU \citep{zhu2018selfbleu} measures the lexical diversity of a text corpus. It leverages the BLEU metric \citep{papieni2002bleu} to evaluate how one sentence
resembles the rest in a collection.
It is obtained by averaging the BLEU score of each sentence of the corpus (hypothesis) with the rest of the collection (reference).
A sample of the corpus is typically used as the computation is too resource-intensive otherwise; we use a sample size of $5,000$ test questions per dataset.

\paragraph{ANLS.} The Average Normalised Levenshtein Similarity is a string edit distance that measures the number of edits it takes to transform one string into another, normalized against string length.
It is defined as
\begin{equation*}
    \mathrm{ANLS}(s, t) \defeq 
    \begin{cases}
        1 - \mathrm{NL}(s, t) & \textbf{if } \mathrm{NL}(s, t) < \tau  \\
        0 & \textbf{otherwise}
    \end{cases}
\end{equation*}
where $\mathrm{NL}(s, t)$ is the normalized Levenshtein Distance between $s$ and $t$, and $\tau$ is a a distance threshold (typically $0.5$).
The reported score over a dataset is the mean ANLS score over all answers in a dataset.

%% file: appendices/a4furtherexperiments.tex
\section{Additional Experimental Results}\label{app:results}
In this section, we provide the results of additional experiments that we could not provide in the main paper due to space constraints.

\subsection{Model Performance on \simpleDatasetName}\label{app:results:simple}
\cref{tab:simple-results} gives the ANLS scores of our different modeling set-ups on the \simpleDatasetName test set.
As one would expect, training on \simpleDatasetName yields the best outcomes.
However, as observed in \cref{sec:experiments:results}, DocLLM performs impressively when fine-tuned using the more diverse templates of \datasetName.
This suggests that the jump in model size from Donut and Pix2Struct to DocLLM enables better generalization with high-quality data.

\subsection{Ablation Studies}\label{app:results:ablation}
\datasetName contains several interesting properties such as the existence of extractive and boolean questions and questions that contain multiple entities.
We conduct an ablation study on how different training configurations and different models perform different question types and number of entities.
Number of entities is counted as number of entities used within the question.
Therefore, the minimum number of entities for extractive questions is zero while the minimum number of entities for boolean questions is one.
These are given in \cref{tab:ablation-ad-buy}, \cref{tab:ablation-cord}, \cref{tab:ablation-docile}, \cref{tab:ablation-klc}, and \cref{tab:ablation-reg-form}. Observe that Pix2Struct and Donut are unable to answer any boolean questions under the zero-shot setting or when trained on \simpleDatasetName.
This demonstrates the difficulty of generalizing beyond the types of questions seen during training. In general, the performance of the model decreases as the number of entities in the question increases.

\input{tables/simple_results}

\input{tables/ablation_ad_buy}
\input{tables/ablation_cord}
\input{tables/ablation_docile}
\input{tables/ablation_klc}
\input{tables/ablation_reg_form}

\subsection{Extended Post-training Assessment of Model Robustness to Template Change}\label{app:results:templates}
We give more results for the experiments described in \cref{sec:experiments:deltas}. \cref{fig:template-deltas-full} gives a complementary figure to \cref{fig:template-deltas} but using \datasetName rather than \sampledDatasetName with similar patterns observed.
Since both \cref{fig:template-deltas} and \cref{fig:template-deltas-full} display relative differences, we illustrate the ANLS scores used to derive them in \cref{fig:template-performance} and \cref{fig:template-performance-full}.

\input{figures/template_deltas_full}
\input{figures/template_performance}
\input{figures/template_performance_full}

\subsection{Extended Evaluation of Generated Errors}\label{app:results:error}
We give a breakdown of the groundedness and types of errors exhibited by our fine-tuned models on Ad-Buy, CORD, Docile, and Reg. Form in \cref{fig:grounded:ad-buy}, \cref{fig:grounded:cord}, \cref{fig:grounded:docile}, and \cref{fig:grounded:reg-form} respectively.
We note that using \datasetName does not always yield better groundedness than \simpleDatasetName.
This is likely due to the clarity of simple templates asking for an entity, specifically for CORD and Docile, for which there are many different entities.
Nevertheless, we observe that for the majority of datasets, training using a version of \datasetName 
leads to higher groundedness as well as a higher number of misprints.
Misprints are a good error to observe as it indicates that the model was able to understand the question and answer, but could not generate the exact text.

We provide a few concrete examples of errors observed in Reg. Form below.

\input{figures/groundedness/grounded_ad_buy}
\input{figures/groundedness/grounded_cord}
\input{figures/groundedness/grounded_docile}
\input{figures/groundedness/grounded_reg_form}

\subsubsection{Error Examples.}\label{app:results:error_examples}
We analyze some of the errors made by DocLLM (the best-performing model) on Reg. Form as a guide for future model development. 
Given that \datasetName closely resembles the kind of questions users in the wild are likely to ask, such an error analysis provides insights as to what errors downstream applications can expect.
An analysis using \simpleDatasetName would not be able to provide such insights.

A representative example of an incorrect answer is from a question ``Who is the signer of the form?'' in which the OCR output for one of the documents is ``/ Chad Horrell'' while the generated answer is ``Chad Horrell''.
We noticed incorrect answers are often subsequences of the ground truth answer or token sequences with a slight offset from the ground truth.
In the above example, the solution may be to start with higher quality annotations as the ``/'' seems out of place.
Another example of what seems to be a misprint error is the question ``What is the registration ID of this form?'' which produces the answer ``6228'' instead of the ground truth ``6278''. 

We also observe many errors that seem to be unrelated to the question or document.
For instance, DocLLM generated the answer ``Government Of Japan--Japan External Trade Organization'' instead of ``Embassy Of The State Of Qatar'' for the question, ``Which foreign principal is this form about?''.
The generated answer is not present in the document but is present in a different document of the training dataset.
This could suggest that the models may suffer from overfitting or hallucination issues. 

We note also that numbers and dates can be reformatted in the generated output which can lead to ungrounded outputs with low ANLS. We evaluated this case and found that this type of error occurs in less than $0.5\%$ of cases across datasets and models in the fine-tuned setting.
Therefore, we do not categorize such errors here and place them in \defn{other}.

%% file: tables/simple_results.tex
\begin{table}[t]
    \small
    \centering
    \begin{tabular}{L{1.2cm} R{0.7cm} R{.95cm} R{.85cm} R{.7cm} R{0.7cm}}
         \bf Model & \bf Ad-Buy & \bf CORD & \bf Docile & \bf KLC & \bf Reg. Form \\ \midrule
         Dnt$_{\text{ZS}}$ & $4.8$ & $61.1$ & $8.9$ & $7.6$ & $15.5$ \\
         P2S$_{\text{ZS}}$ & $19.1$ & $63.8$ & $25.2$ & $12.8$ & $30.0$ \\
        Doc$_{\text{ZS}}$ & $1.6$ & $72.9$ & $28.1$ & $\boldsymbol{78.2}$ & $2.8$ \\
      UDOP$_{\text{ZS}}$ & $4.9$ & $44.2$ & $21.6$ & $12.8$ & $4.8$ \\

         mPD$_{\text{ZS}}$ & $64.4$ & $55.2$ & $41.8$ & $71.6$ & $60.5$ \\
         mPDC$_{\text{ZS}}$ & $\boldsymbol{65.9}$ & $42.5$ & $41.2$ & $71.8$ & $59.8$ \\
         GPT-4$_{\text{ZS}}$ & $50.5$ & $\boldsymbol{82.4}$ & $\boldsymbol{48.0}$ & $38.7$ & $\boldsymbol{70.2}$ \\
         \midrule
         Dnt$_{\text{\simpleDatasetName}}$ & $61.2$ & $90.2$ & $39.9$ & $55.9$ & $78.3$ \\
         P2S$_{\text{\simpleDatasetName}}$ & $79.0$ & $89.3$ & $61.2$ & $76.3$ & $\boldsymbol{85.6}$ \\
         Doc$_{\text{\simpleDatasetName}}$ & $\boldsymbol{95.7}$ & $\boldsymbol{94.4}$ & $\boldsymbol{86.7}$ & $\boldsymbol{88.9}$ & $66.7$ \\
         
         \midrule
         Dnt$_{\text{\sampledDatasetName}}$ & $24.0$ & $80.3$ & $19.2$ & $24.1$ & $38.8$ \\
         P2S$_{\text{\sampledDatasetName}}$ & $27.7$ & $81.7$ & $27.6$ & $29.0$ & $33.9$ \\
Doc$_{\text{\sampledDatasetName}}$ & $\boldsymbol{91.1}$ & $\boldsymbol{94.3}$ & $\boldsymbol{82.1}$ & $\boldsymbol{85.6}$ & $\boldsymbol{62.6}$ \\
         \midrule
         Dnt$_{\text{\datasetName}}$ & $24.2$ & $81.7$ & $19.3$ & $24.3$ & $35.9$ \\
         P2S$_{\text{\datasetName}}$ & $28.2$ & $81.6$ & $27.6$ & $29.9$ & $28.2$ \\
         Doc$_{\text{\datasetName}}$ & $\boldsymbol{95.2}$ & $\boldsymbol{95.9}$ & $\boldsymbol{82.5}$ & $\boldsymbol{87.2}$ & $\boldsymbol{66.9}$ \\
    \end{tabular}
    \caption{ANLS ($\uparrow$) results on \simpleDatasetName test set using various training settings for models Donut (Dnt), Pix2Struct (P2S), DocLLM (Doc), UDOP, mPlugDocOwl 1.5 (mPD), mPlugDocOwl 1.5-Chat (mPDC), and GPT-4.
    }
    \label{tab:simple-results}
\end{table}

%% file: tables/ablation_ad_buy.tex
\begin{table*}[t]
    \centering
    \small
    \begin{tabular}{l r rrrrr rrr}
        \multirow{2}{*}{\bf Model} & \multirow{2}{*}{\bf All} & \multicolumn{5}{c}{\bf Extractive Questions} & \multicolumn{3}{c}{\bf Boolean Questions} \\
        \cmidrule(lr){3-7}
        \cmidrule(lr){8-10}
        & &
        \multicolumn{1}{c}{\bf All} &
        \multicolumn{1}{c}{\bf 0 Entities} & 
        \multicolumn{1}{c}{\bf 1 Entity} & 
        \multicolumn{1}{c}{\bf 2 Entities} & 
        \multicolumn{1}{c}{\bf 3 Entities} & 
        \multicolumn{1}{c}{\bf All} &
        \multicolumn{1}{c}{\bf 1 Entity} & 
        \multicolumn{1}{c}{\bf 2 Entities} \\ \midrule
Donut$_{\text{ZS}}$ & $1.4$ & $2.5$ & $2.9$ & $1.7$ & $0.6$ & $19.0$ & $0.0$ & $0.0$ & $0.0$ \\
Pix2Struct$_{\text{ZS}}$ & $13.5$ & $24.1$ & $26.2$ & $15.9$ & $26.4$ & $32.6$ & $0.0$ & $0.0$ & $0.0$ \\
DocLLM$_{\text{ZS}}$ & $23.9$ & $5.3$ & $5.6$ & $2.2$ & $7.4$ & $16.8$ & $47.6$ & $47.3$ & $48.1$ \\
UDOP$_{\text{ZS}}$ & $29.1$ & $22.0$ & $16.4$ & $23.4$ & $36.1$ & $42.4$ & $38.2$ & $42.8$ & $27.7$ \\
mPlugDO$_{\text{ZS}}$ & $60.9$ & $47.2$ & $52.2$ & $31.0$ & $54.0$ & $25.8$ & $78.4$ & $82.2$ & $69.6$ \\
mPlugDOC$_{\text{ZS}}$ & $61.9$ & $53.1$ & $57.0$ & $38.9$ & $59.2$ & $44.2$ & $73.2$ & $76.4$ & $65.9$ \\
GPT-4$_{\text{ZS}}$ & $72.7$ & $68.1$ & $65.2$ & $65.6$ & $79.6$ & $76.7$ & $78.6$ & $76.0$ & $84.5$ \\
\midrule
Donut$_{\text{\simpleDatasetName}}$ & $7.7$ & $13.7$ & $18.9$ & $6.9$ & $5.7$ & $7.5$ & $0.0$ & $0.0$ & $0.0$ \\
Pix2Struct$_{\text{\simpleDatasetName}}$ & $18.0$ & $32.2$ & $37.4$ & $19.7$ & $32.9$ & $13.3$ & $0.0$ & $0.0$ & $0.0$ \\
DocLLM$_{\text{\simpleDatasetName}}$ & $46.3$ & $52.2$ & $60.2$ & $35.7$ & $45.8$ & $61.5$ & $38.7$ & $38.9$ & $38.2$ \\ \midrule
Donut$_{\text{\sampledDatasetName}}$ & $56.4$ & $48.2$ & $44.6$ & $44.6$ & $62.8$ & $65.1$ & $66.9$ & $69.7$ & $60.5$ \\
Pix2Struct$_{\text{\sampledDatasetName}}$ & $69.2$ & $64.5$ & $64.3$ & $61.7$ & $68.6$ & $65.8$ & $75.2$ & $79.0$ & $66.5$ \\
DocLLM$_{\text{\sampledDatasetName}}$ & $92.6$ & $89.3$ & $93.6$ & $83.1$ & $84.2$ & $77.5$ & $96.7$ & $97.6$ & $94.8$ \\ \midrule
Donut$_{\text{\datasetName}}$ & $58.5$ & $49.1$ & $45.9$ & $44.8$ & $63.4$ & $63.1$ & $70.5$ & $73.9$ & $62.6$ \\
Pix2Struct$_{\text{\datasetName}}$ & $73.8$ & $68.6$ & $69.6$ & $63.2$ & $72.1$ & $69.2$ & $80.3$ & $84.7$ & $70.2$ \\
DocLLM$_{\text{\datasetName}}$ & $93.9$ & $91.5$ & $96.2$ & $85.6$ & $85.3$ & $77.4$ & $96.9$ & $97.6$ & $95.2$ \\
    \end{tabular}
    \caption{Ablation study of ANLS ($\uparrow$) for \datasetName Ad-Buy.}
    \label{tab:ablation-ad-buy}
\end{table*}

%% file: tables/ablation_cord.tex
\begin{table*}[t]
    \centering
    \small
    \begin{tabular}{l r rrr r}
        \multirow{2}{*}{\bf Model} & \multirow{2}{*}{\bf All} & \multicolumn{3}{c}{\bf Extractive Questions} & \multirow{2}{2cm}{\makecell{\bf Boolean \\ \bf Questions}} \\
        \cmidrule(lr){3-5}
        & &
        \multicolumn{1}{c}{\bf All} &
        \multicolumn{1}{c}{\bf 0 Entities} & 
        \multicolumn{1}{c}{\bf 1 Entity} & 
         \\ \midrule
Donut$_{\text{ZS}}$ & $18.0$ & $36.3$ & $62.1$ & $26.9$ & $0.0$  \\
Pix2Struct$_{\text{ZS}}$ & $27.0$ & $54.3$ & $68.7$ & $49.1$ & $0.0$  \\
DocLLM$_{\text{ZS}}$ & $43.0$ & $62.9$ & $83.7$ & $55.4$ & $23.4$  \\
UDOP$_{\text{ZS}}$ & $29.7$ & $33.0$ & $49.0$ & $27.2$ & $26.4$  \\
mPlugDO$_{\text{ZS}}$ & $64.5$ & $76.8$ & $72.9$ & $78.2$ & $52.3$  \\
mPlugDOC$_{\text{ZS}}$ & $64.4$ & $76.2$ & $74.0$ & $77.0$ & $52.8$  \\
GPT-4$_{\text{ZS}}$ & $85.3$ & $87.7$ & $88.0$ & $87.6$ & $82.9$  \\ \midrule
Donut$_{\text{\simpleDatasetName}}$ & $24.4$ & $49.0$ & $86.1$ & $35.5$ & $0.0$  \\
Pix2Struct$_{\text{\simpleDatasetName}}$ & $31.0$ & $62.4$ & $89.3$ & $52.6$ & $0.0$  \\
DocLLM$_{\text{\simpleDatasetName}}$ & $43.6$ & $70.8$ & $95.6$ & $61.9$ & $16.6$  \\ \midrule
Donut$_{\text{\sampledDatasetName}}$ & $82.7$ & $83.8$ & $84.2$ & $83.6$ & $81.7$  \\
Pix2Struct$_{\text{\sampledDatasetName}}$ & $84.7$ & $88.4$ & $86.6$ & $89.0$ & $81.1$  \\
DocLLM$_{\text{\sampledDatasetName}}$ & $94.0$ & $95.1$ & $96.8$ & $94.5$ & $93.0$  \\ \midrule
Donut$_{\text{\datasetName}}$ & $83.3$ & $85.1$ & $84.3$ & $85.4$ & $81.5$  \\
Pix2Struct$_{\text{\datasetName}}$ & $86.5$ & $90.2$ & $88.8$ & $90.6$ & $83.0$  \\
DocLLM$_{\text{\datasetName}}$ & $96.5$ & $96.6$ & $96.9$ & $96.5$ & $96.3$  \\
    \end{tabular}
    \caption{Ablation study of ANLS ($\uparrow$) for \datasetName CORD.}
    \label{tab:ablation-cord}
\end{table*}

%% file: tables/ablation_docile.tex
\begin{table*}[t]
    \centering
    \small
    \begin{tabular}{l r rrr r}
        \multirow{2}{*}{\bf Model} & \multirow{2}{*}{\bf All} & \multicolumn{3}{c}{\bf Extractive Questions} & \multirow{2}{2cm}{\makecell{\bf Boolean \\ \bf Questions}} \\
        \cmidrule(lr){3-5}
        & &
        \multicolumn{1}{c}{\bf All} &
        \multicolumn{1}{c}{\bf 0 Entities} & 
        \multicolumn{1}{c}{\bf 1 Entity} & 
         \\ \midrule
Donut$_{\text{ZS}}$ & $7.2$ & $10.4$ & $11.2$ & $0.2$ & $0.0$  \\
Pix2Struct$_{\text{ZS}}$ & $24.1$ & $35.0$ & $37.0$ & $10.1$ & $0.0$  \\
DocLLM$_{\text{ZS}}$ & $48.0$ & $42.1$ & $43.5$ & $23.9$ & $61.0$  \\
UDOP$_{\text{ZS}}$ & $35.1$ & $33.2$ & $34.7$ & $14.0$ & $39.2$  \\
mPlugDO$_{\text{ZS}}$ & $56.0$ & $49.7$ & $52.2$ & $18.0$ & $69.7$  \\
mPlugDOC$_{\text{ZS}}$ & $53.4$ & $50.4$ & $53.0$ & $17.5$ & $59.9$  \\
GPT-4$_{\text{ZS}}$ & $61.2$ & $54.0$ & $54.5$ & $47.0$ & $77.3$  \\ \midrule
Donut$_{\text{\simpleDatasetName}}$ & $22.3$ & $32.4$ & $34.5$ & $5.2$ & $0.0$  \\
Pix2Struct$_{\text{\simpleDatasetName}}$ & $35.5$ & $51.6$ & $53.0$ & $33.1$ & $0.0$  \\
DocLLM$_{\text{\simpleDatasetName}}$ & $53.0$ & $76.4$ & $76.9$ & $70.0$ & $1.5$  \\ \midrule
Donut$_{\text{\sampledDatasetName}}$ & $47.7$ & $39.0$ & $40.9$ & $14.4$ & $66.9$  \\
Pix2Struct$_{\text{\sampledDatasetName}}$ & $59.0$ & $60.1$ & $60.7$ & $51.8$ & $56.7$  \\
DocLLM$_{\text{\sampledDatasetName}}$ & $88.9$ & $85.7$ & $85.8$ & $84.5$ & $95.8$  \\ \midrule
Donut$_{\text{\datasetName}}$ & $47.8$ & $39.1$ & $41.0$ & $14.4$ & $67.2$  \\
Pix2Struct$_{\text{\datasetName}}$ & $59.5$ & $60.6$ & $61.2$ & $53.4$ & $57.1$  \\
DocLLM$_{\text{\datasetName}}$ & $90.0$ & $87.0$ & $87.2$ & $83.7$ & $96.8$  \\
    \end{tabular}
    \caption{Ablation study of ANLS ($\uparrow$) for \datasetName Docile.}
    \label{tab:ablation-docile}
\end{table*}

%% file: tables/ablation_klc.tex
\begin{table*}[t]
    \centering
    \small
    \begin{tabular}{l r rrr rrr}
        \multirow{2}{*}{\bf Model} & \multirow{2}{*}{\bf All} & \multicolumn{3}{c}{\bf Extractive Questions} & \multicolumn{3}{c}{\bf Boolean Questions} \\
        \cmidrule(lr){3-5}
        \cmidrule(lr){6-8}
        & &
        \multicolumn{1}{c}{\bf All} &
        \multicolumn{1}{c}{\bf 0 Entities} & 
        \multicolumn{1}{c}{\bf 1 Entity} & 
        \multicolumn{1}{c}{\bf All} &
        \multicolumn{1}{c}{\bf 1 Entity} & 
        \multicolumn{1}{c}{\bf 2 Entities} \\ \midrule
Donut$_{\text{ZS}}$ & $3.8$ & $9.2$ & $19.9$ & $5.1$ & $0.1$ & $0.3$ & $0.1$ \\
Pix2Struct$_{\text{ZS}}$ & $11.8$ & $28.8$ & $38.7$ & $25.0$ & $0.0$ & $0.0$ & $0.0$ \\
DocLLM$_{\text{ZS}}$ & $80.4$ & $80.4$ & $85.1$ & $78.7$ & $80.4$ & $80.1$ & $80.5$ \\
UDOP$_{\text{ZS}}$ & $30.0$ & $22.2$ & $31.5$ & $18.7$ & $35.4$ & $50.7$ & $30.6$ \\
mPlugDO$_{\text{ZS}}$ & $66.8$ & $65.6$ & $66.7$ & $65.2$ & $67.6$ & $78.4$ & $64.2$ \\
mPlugDOC$_{\text{ZS}}$ & $66.0$ & $64.9$ & $64.1$ & $65.3$ & $66.7$ & $81.0$ & $62.4$ \\
GPT-4$_{\text{ZS}}$ & $68.1$ & $65.7$ & $53.4$ & $70.3$ & $69.8$ & $62.1$ & $72.2$ \\ \midrule
Donut$_{\text{\simpleDatasetName}}$ & $15.6$ & $37.9$ & $66.7$ & $27.0$ & $0.0$ & $0.1$ & $0.0$ \\
Pix2Struct$_{\text{\simpleDatasetName}}$ & $28.0$ & $68.4$ & $89.2$ & $60.5$ & $0.0$ & $0.0$ & $0.0$ \\
DocLLM$_{\text{\simpleDatasetName}}$ & $62.8$ & $88.7$ & $95.8$ & $86.0$ & $44.8$ & $47.2$ & $44.0$ \\ \midrule
Donut$_{\text{\sampledDatasetName}}$ & $67.2$ & $45.9$ & $75.6$ & $34.6$ & $82.1$ & $86.8$ & $80.6$ \\
Pix2Struct$_{\text{\sampledDatasetName}}$ & $77.0$ & $72.4$ & $90.7$ & $65.5$ & $80.2$ & $81.3$ & $79.9$ \\
DocLLM$_{\text{\sampledDatasetName}}$ & $92.6$ & $88.4$ & $94.3$ & $86.2$ & $95.5$ & $98.2$ & $94.7$ \\ \midrule
Donut$_{\text{\datasetName}}$ & $68.1$ & $46.1$ & $76.0$ & $34.8$ & $83.3$ & $87.8$ & $82.0$ \\
Pix2Struct$_{\text{\datasetName}}$ & $79.4$ & $74.0$ & $91.5$ & $67.3$ & $83.2$ & $84.1$ & $82.9$ \\
DocLLM$_{\text{\datasetName}}$ & $93.6$ & $89.7$ & $95.8$ & $87.4$ & $96.3$ & $98.9$ & $95.4$ \\
    \end{tabular}
    \caption{Ablation study of ANLS ($\uparrow$) for \datasetName KLC.}
    \label{tab:ablation-klc}
\end{table*}

%% file: tables/ablation_reg_form.tex
\begin{table*}[t]
    \centering
    \small
    \begin{tabular}{l r rrrrr rrr}
        \multirow{2}{*}{\bf Model} & \multirow{2}{*}{\bf All} & \multicolumn{3}{c}{\bf Extractive Questions} & \multicolumn{3}{c}{\bf Boolean Questions} \\
        \cmidrule(lr){3-5}
        \cmidrule(lr){6-8}
        & &
        \multicolumn{1}{c}{\bf All} &
        \multicolumn{1}{c}{\bf 0 Entities} & 
        \multicolumn{1}{c}{\bf 1 Entity} & 
        \multicolumn{1}{c}{\bf All} &
        \multicolumn{1}{c}{\bf 1 Entity} & 
        \multicolumn{1}{c}{\bf 2 Entities} \\ \midrule
Donut$_{\text{ZS}}$ & $5.9$ & $15.8$ & $15.2$ & $25.0$ & $0.0$ & $0.0$ & $0.0$ \\
Pix2Struct$_{\text{ZS}}$ & $24.0$ & $63.7$ & $62.1$ & $91.3$ & $0.0$ & $0.0$ & $0.0$ \\
DocLLM$_{\text{ZS}}$ & $27.6$ & $4.2$ & $4.1$ & $5.0$ & $41.7$ & $41.8$ & $40.3$ \\
UDOP$_{\text{ZS}}$ & $39.1$ & $19.0$ & $16.9$ & $53.5$ & $51.2$ & $51.6$ & $45.3$ \\
mPlugDO$_{\text{ZS}}$ & $65.2$ & $62.0$ & $60.1$ & $93.0$ & $67.1$ & $66.3$ & $78.8$ \\
mPlugDOC$_{\text{ZS}}$ & $66.8$ & $65.5$ & $63.9$ & $92.2$ & $67.6$ & $67.4$ & $71.2$ \\
GPT-4$_{\text{ZS}}$ & $76.5$ & $74.6$ & $73.2$ & $97.1$ & $77.7$ & $78.4$ & $66.1$ \\ \midrule
Donut$_{\text{\simpleDatasetName}}$ & $19.7$ & $52.4$ & $51.0$ & $75.3$ & $0.0$ & $0.0$ & $0.0$ \\
Pix2Struct$_{\text{\simpleDatasetName}}$ & $29.7$ & $78.8$ & $78.3$ & $87.0$ & $0.0$ & $0.0$ & $0.0$ \\
DocLLM$_{\text{\simpleDatasetName}}$ & $66.2$ & $64.2$ & $62.6$ & $91.1$ & $67.5$ & $68.1$ & $58.5$ \\ \midrule
Donut$_{\text{\sampledDatasetName}}$ & $81.4$ & $69.7$ & $69.6$ & $71.5$ & $88.5$ & $89.3$ & $77.1$ \\
Pix2Struct$_{\text{\sampledDatasetName}}$ & $87.0$ & $81.1$ & $80.9$ & $84.3$ & $90.7$ & $91.1$ & $84.7$ \\
DocLLM$_{\text{\sampledDatasetName}}$ & $78.0$ & $62.3$ & $60.7$ & $88.5$ & $87.4$ & $87.1$ & $91.9$ \\ \midrule
Donut$_{\text{\datasetName}}$ & $82.7$ & $70.7$ & $70.8$ & $69.3$ & $90.0$ & $90.4$ & $83.1$ \\
Pix2Struct$_{\text{\datasetName}}$ & $88.7$ & $82.7$ & $82.7$ & $82.5$ & $92.3$ & $92.5$ & $89.4$ \\
DocLLM$_{\text{\datasetName}}$ & $80.3$ & $66.2$ & $64.7$ & $91.0$ & $88.9$ & $88.4$ & $95.8$ \\
    \end{tabular}
    \caption{Ablation study of ANLS ($\uparrow$) for \datasetName Reg. Form.}
    \label{tab:ablation-reg-form}
\end{table*}

%% file: figures/template_deltas_full.tex
\begin{figure*}[t]
    \centering     \includegraphics[width=\textwidth]{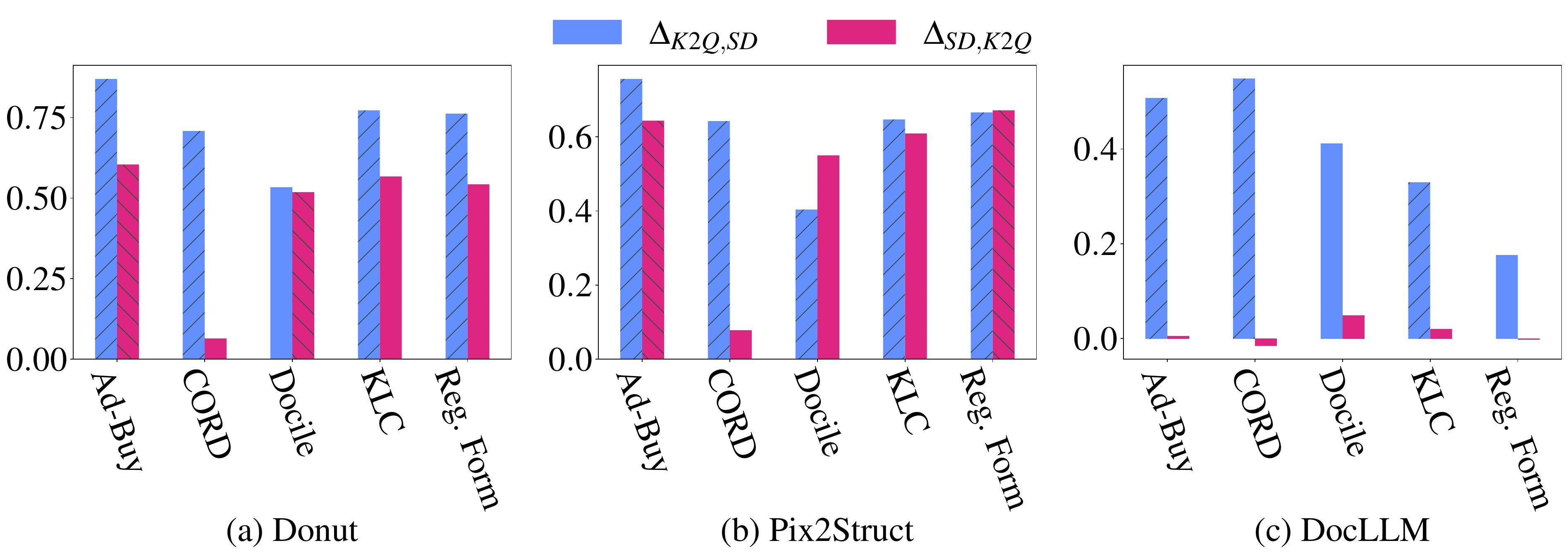}
    \caption{Comparison of training and evaluating on complex questions (\datasetName) and simple questions (\simpleDatasetName).}
    \label{fig:template-deltas-full}
\end{figure*}

%% file: figures/template_performance.tex
\begin{figure*}[t]
    \centering    \includegraphics[width=\textwidth]{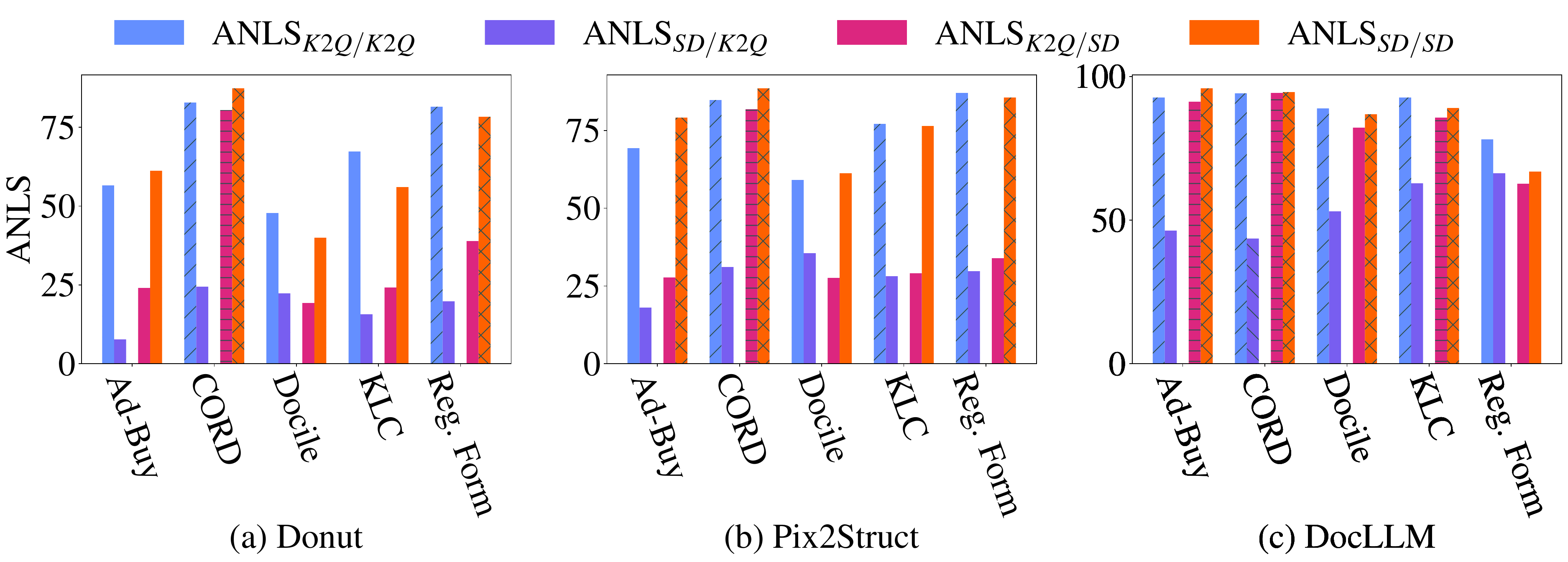}
    \caption{Comparison of ANLS scores for training and evaluating on sampled complex questions (\sampledDatasetName) and simple questions (\simpleDatasetName). }
    \label{fig:template-performance}
\end{figure*}

%% file: figures/template_performance_full.tex
\begin{figure*}[t]
    \centering    \includegraphics[width=\textwidth]{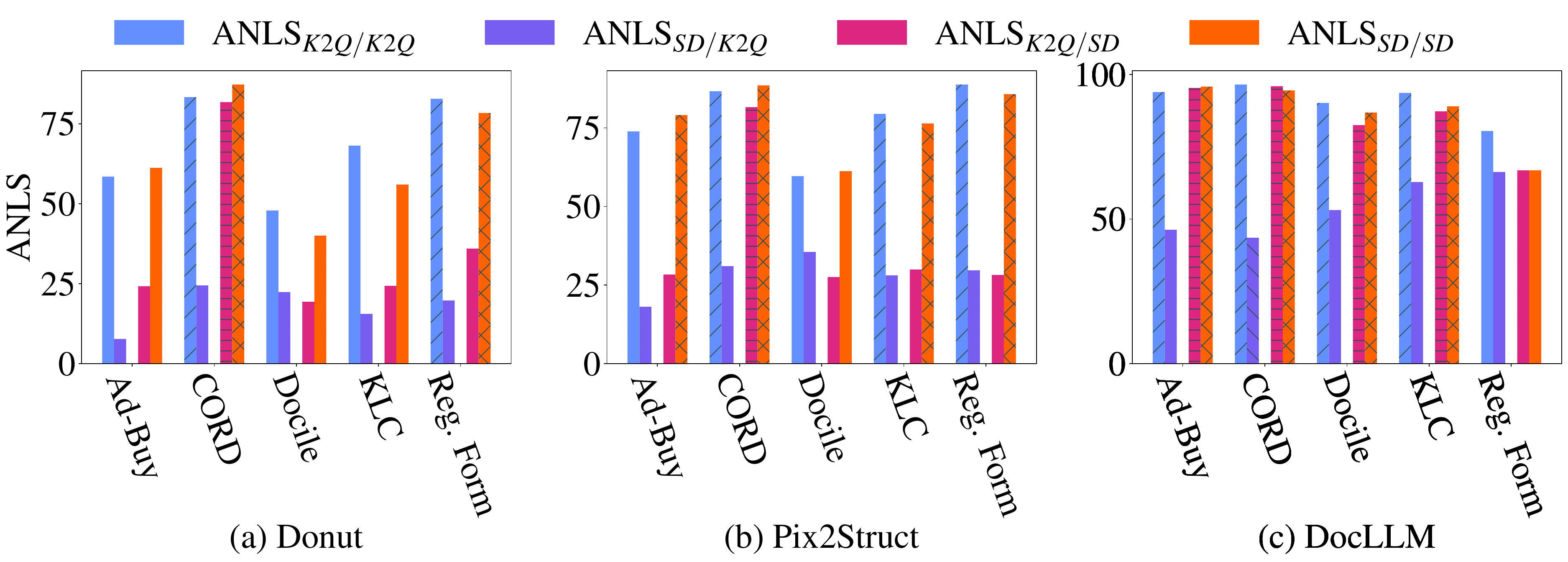}
    \caption{Comparison of ANLS scores for training and evaluating on complex questions (\datasetName) and simple questions (\simpleDatasetName). }
    \label{fig:template-performance-full}
\end{figure*}

%% file: figures/groundedness/grounded_ad_buy.tex
\begin{figure*}[t]
    \centering     \includegraphics[width=\textwidth]{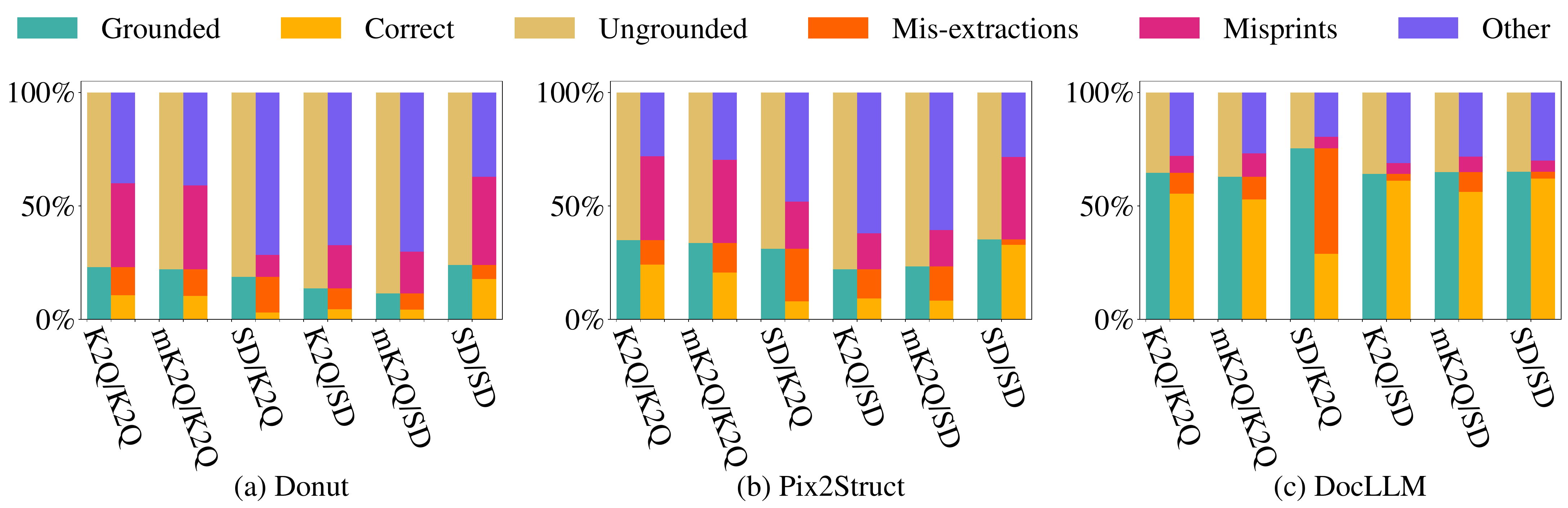}
    \caption{Detailed breakdown of groundedness and error types for Ad-Buy using different training / testing datasets.}
    \label{fig:grounded:ad-buy}
\end{figure*}

%% file: figures/groundedness/grounded_cord.tex
\begin{figure*}[t]
    \centering     \includegraphics[width=\textwidth]{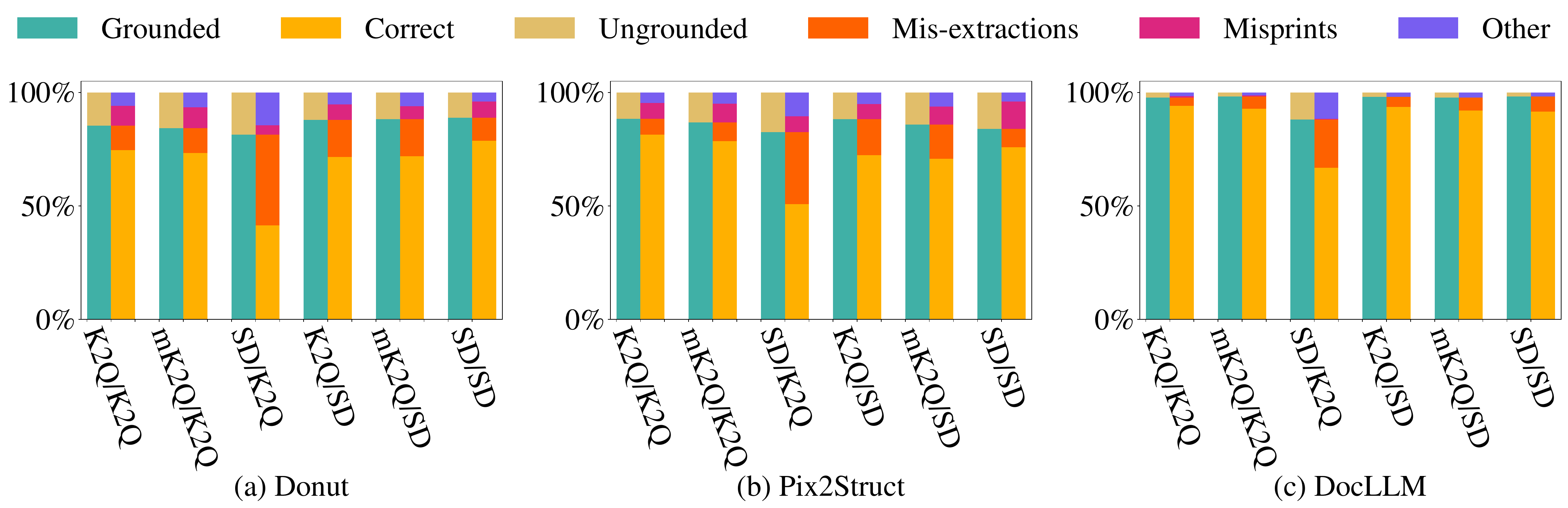}
    \caption{Detailed breakdown of groundedness and error types for CORD using different training / testing datasets.}
    \label{fig:grounded:cord}
\end{figure*}

%% file: figures/groundedness/grounded_docile.tex
\begin{figure*}[t]
    \centering     \includegraphics[width=\textwidth]{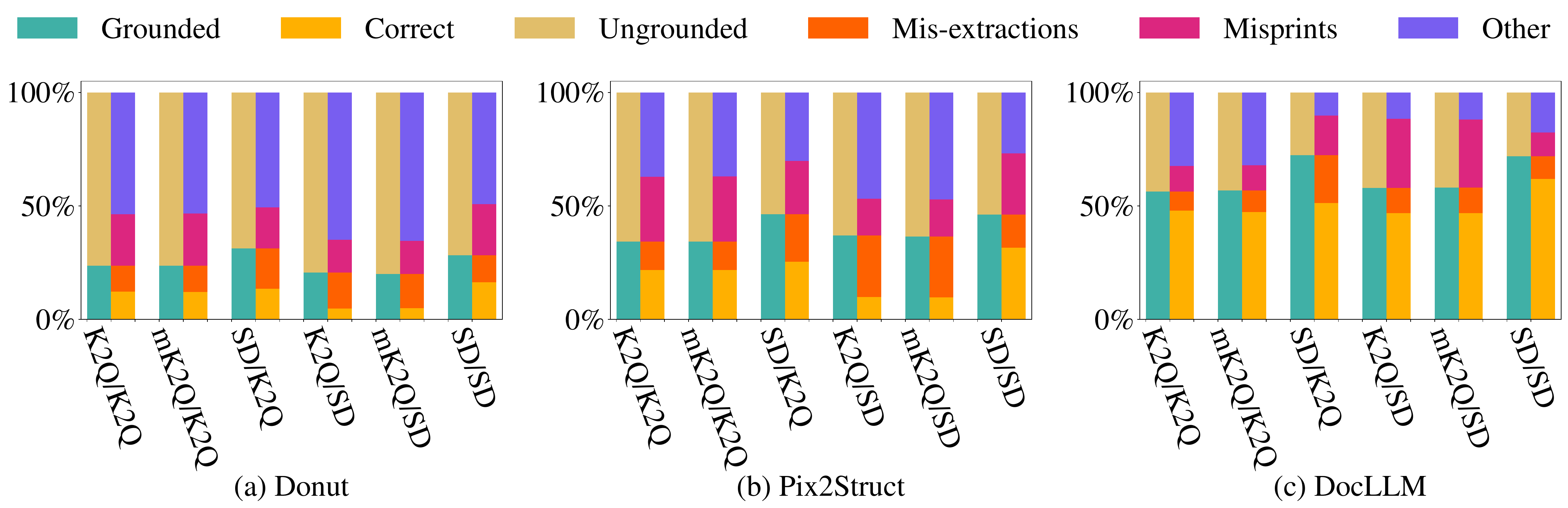}
    \caption{Detailed breakdown of groundedness and error types for Docile using different training / testing datasets.}
    \label{fig:grounded:docile}
\end{figure*}

%% file: figures/groundedness/grounded_reg_form.tex
\begin{figure*}[t]
    \centering     \includegraphics[width=\textwidth]{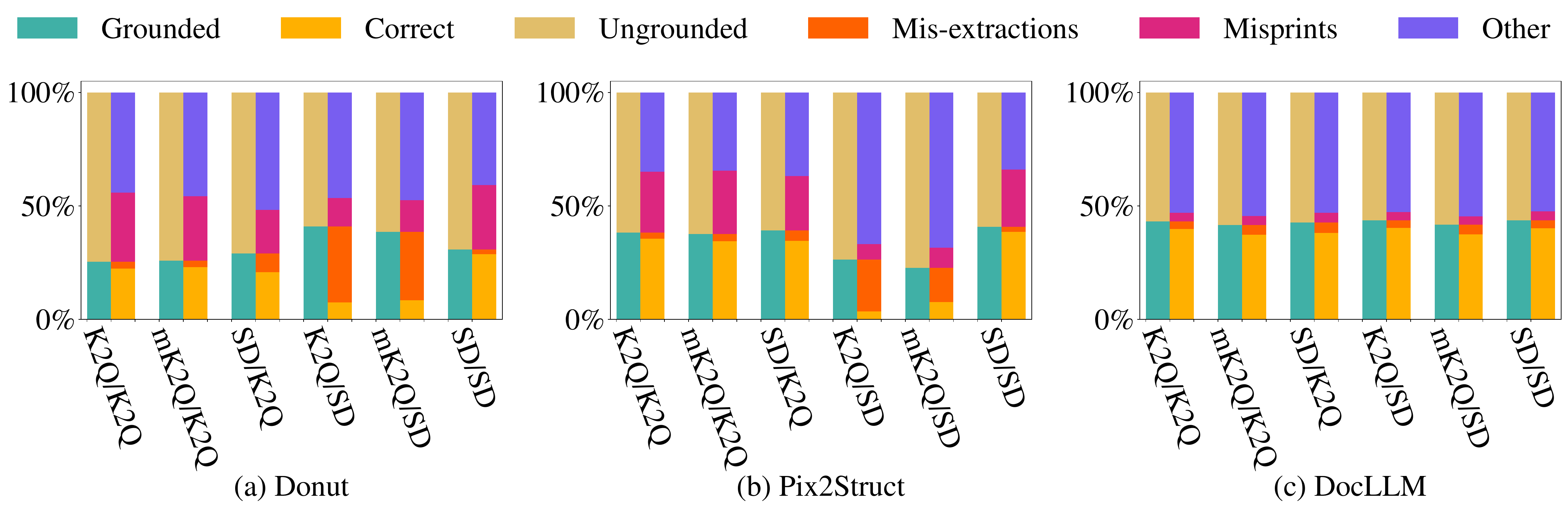}
    \caption{Detailed breakdown of groundedness and error types for Reg. Form using different training / testing datasets.}
    \label{fig:grounded:reg-form}
\end{figure*}